\PassOptionsToPackage{hyperfootnotes=false}{hyperref}
\RequirePackage{fix-cm}
\documentclass[11pt,a4paper,onecolumn]{personalarxiv}
\usepackage[UTF8,fontset=fandol,scheme=plain]{ctex}
\microtypesetup{tracking=false}
\usepackage{mathtools,array,longtable}
\usepackage[section]{placeins}
\usepackage{fontawesome5}
\usepackage{xcolor}
\usepackage{colortbl}
\usepackage{tikz}
\usepackage{hyperref}

\newcolumntype{Y}{>{\raggedright\arraybackslash}X}
\newcolumntype{P}[1]{>{\raggedright\arraybackslash}p{#1}}
\newcommand{\E}{\mathbb E}
\newcommand{\KL}{D_{\mathrm{KL}}}

\newcommand{\GenPPL}{\operatorname{GenPPL}}
\newcommand{\Hrow}{H_{\mathrm{row}}}

\newcommand{\JSDrows}{\operatorname{JSD}_{\mathrm{rows}}}
\newcommand{\bench}{\textsc{Open CDLM}}
\newtheorem{proposition}{Proposition}[section]
\setlist[itemize]{topsep=3pt,itemsep=3pt}
\setlist[enumerate]{topsep=3pt,itemsep=3pt}
\renewcommand{\headerfont}{\normalfont\color{black}\fontsize{8}{10}\selectfont}

\definecolor{CDLMnoise}{HTML}{22BCD0}
\definecolor{CDLMflow}{HTML}{4879ED}
\definecolor{CDLMtoken}{HTML}{7B4FE0}

\DeclareRobustCommand{\CDLMicon}{%
  \begin{tikzpicture}[
    x=0.5em,y=0.5em,baseline=-0.22em
  ]
    \shade[left color=CDLMnoise,right color=CDLMtoken]
      (0.61,0.82)
      .. controls (0.18,1.17) and (-0.54,1.08) .. (-0.91,0.63)
      .. controls (-1.37,0.10) and (-1.16,-0.68) .. (-0.55,-0.94)
      .. controls (-0.17,-1.13) and (0.34,-0.98) .. (0.63,-0.68)
      -- (0.39,-0.46)
      .. controls (0.12,-0.71) and (-0.21,-0.77) .. (-0.49,-0.61)
      .. controls (-0.95,-0.38) and (-0.98,0.19) .. (-0.65,0.55)
      .. controls (-0.38,0.88) and (0.09,0.92) .. (0.44,0.67)
      -- cycle;

    \shade[
      left color=CDLMflow,right color=CDLMtoken,
      rounded corners=0.065em
    ] (-0.66,-0.30) rectangle (0.28,0.43);

    \fill[CDLMtoken]
      (0.02,-0.28) -- (0.23,-0.46)
      -- (0.20,-0.23) -- cycle;

    \draw[white,line width=0.038em,line cap=round]
      (-0.47,0.23)  -- (0.08,0.23)
      (-0.47,0.065) -- (0.08,0.065)
      (-0.47,-0.10) -- (-0.12,-0.10);

    \foreach \x/\y/\r/\c/\o in {
      0.48/0.12/0.033/CDLMtoken/1,
      0.54/-0.13/0.030/CDLMtoken/0.95,
      0.64/0.29/0.030/CDLMflow/0.95,
      0.73/0.02/0.032/CDLMflow/0.90,
      0.79/-0.29/0.028/CDLMflow/0.85,
      0.95/0.40/0.027/CDLMflow/0.85,
      1.01/-0.10/0.030/CDLMflow/0.85,
      1.08/0.68/0.023/CDLMnoise/0.70,
      1.20/-0.48/0.027/CDLMflow/0.75,
      1.30/0.17/0.026/CDLMnoise/0.80,
      1.47/0.47/0.024/CDLMnoise/0.75,
      1.50/-0.24/0.025/CDLMnoise/0.75,
      1.70/0.88/0.021/CDLMnoise/0.55,
      1.84/0.08/0.025/CDLMnoise/0.65,
      1.91/-0.70/0.022/CDLMnoise/0.55,
      2.16/0.50/0.020/CDLMnoise/0.50,
      2.29/-0.24/0.021/CDLMnoise/0.50,
      1.54/-0.92/0.020/CDLMnoise/0.45
    }{
      \fill[\c,opacity=\o]
        (\x,\y) circle[radius=\r em];
    }
  \end{tikzpicture}%
}

\title{%
  \texorpdfstring{\CDLMicon\hspace{0.25em}}{}%
  Open CDLM: A Library and Benchmark
  for Continuous Diffusion Language Models
}

\title{%
  Open CDLM: A Benchmark and Library for Continuous Diffusion Language Models
}
\papershorttitle{Open CDLM: Likelihood, Generation Evaluation, and Reproducibility}
\paperdate{}

\author[1,2]{Hao Jiang\thanks{These authors contributed equally.}}
\author[1,2]{Ruixi Chen\protect\footnotemark[1]}
\author[1,2]{Jiayi Han}
\author[1,2]{Yulei Sun}
\author[1,3]{Yujian Chen}
\author[4]{Gen Li}
\author[5]{\protect\\Brian K Chen}
\author[6]{Guang Cheng}
\author[7]{Weijian Luo\thanks{Corresponding author: \href{mailto:pkulwj1994@icloud.com}{\texttt{pkulwj1994@icloud.com}}}}
\author[1,2]{Tianyang Hu\thanks{Corresponding author: \href{mailto:hutianyang@cuhk.edu.cn}{\texttt{hutianyang@cuhk.edu.cn}}}}

\affil[1]{Polyadic Research Group}
\affil[2]{SAGA Lab, The Chinese University of Hong Kong, Shenzhen}
\affil[3]{School of Computer Science, Peking University}
\affil[4]{The Chinese University of Hong Kong}
\affil[5]{National University of Singapore}
\affil[6]{University of California, Los Angeles}
\affil[7]{Dots Studio, Xiaohongshu}

\paperlogos{%
  \resizebox{\textwidth}{!}{%
    \includegraphics[height=10mm]{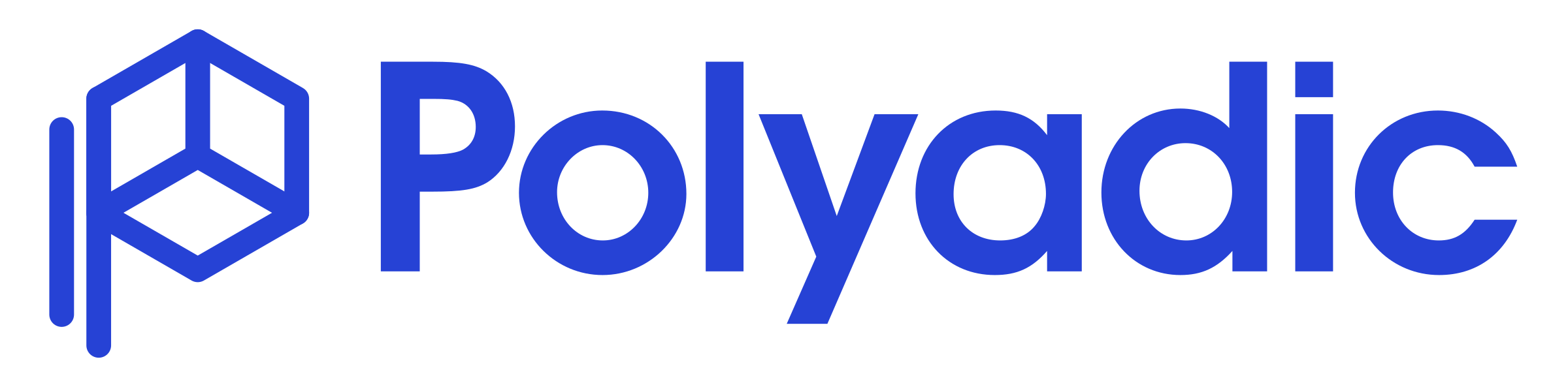}%
    \hspace{7mm}%
    \includegraphics[height=10mm]{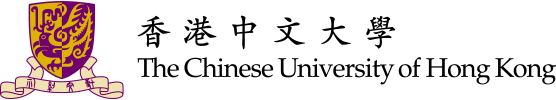}%
    \hspace{7mm}%
    \includegraphics[height=10mm]{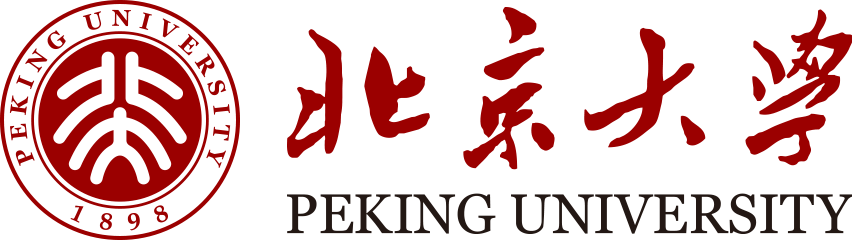}%
    \hspace{7mm}%
    \includegraphics[height=10mm]{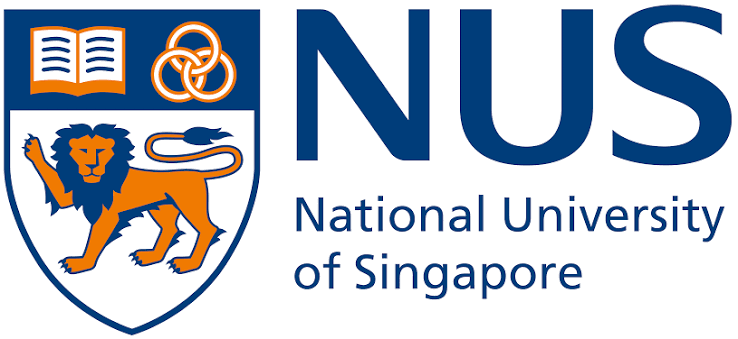}%
    \hspace{7mm}%
    \includegraphics[height=10mm]{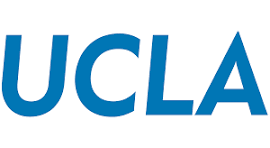}%
  }%
}

\begin{document}
\maketitle
\begin{abstract}
Reliable comparisons of continuous diffusion language models (CDLMs) require evaluating both model fit and the distribution of generated text. 
Two common practices leave this goal only partially addressed. 
Likelihood-based evaluation often reports a variational upper bound on discrete negative log-likelihood, whose slack can confound comparisons. 
Generated-sample-based evaluation commonly combines generative perplexity with within-sequence unigram entropy; both scores can remain favorable even when a sampler produces the same text on every draw. 
We introduce \textbf{Open CDLM}, a benchmark for evaluating CDLMs and their samplers, and describe the design of an accompanying open-source library to support reproducible comparisons. 
For likelihood-based evaluation, we clarify which comparisons of model fit are supported by reported perplexity bounds and how those conclusions depend on the evaluation procedure. 
For sample-based evaluation, we use \textit{Kernel Text Discrepancy} (\textbf{KTD}) to compare collections of generated and real texts, alongside diagnostics for diversity, validity, and coverage.
In controlled OpenWebText experiments, replacing a collection of texts with copies of a single selected passage improves both conventional generation scores. MAUVE detects the resulting collapse, and its response to other perturbations depends on the text representation and the type of perturbation.
This work marks the beginning of Open CDLM as an open-source project for CDLM research. We will maintain and extend the library over time and welcome contributions from the community.

\begin{center}
  \small
  \textcolor[HTML]{6E40C9}{\faGithub}\enspace
  \textbf{Code:}\enspace
  \href{https://github.com/polyadic-research/open_cdlm}
       {\nolinkurl{github.com/polyadic-research/open_cdlm}}
  \\[0.4em]
  \textcolor[HTML]{1A73E8}{\faGlobe}\enspace
  \textbf{Project page:}\enspace
  \href{https://hojawn.github.io/CDLM-Bench/}
       {\nolinkurl{hojawn.github.io/CDLM-Bench/}}
  \\[0.4em]
  \textcolor[HTML]{0F9D8A}{\faUsers}\enspace
  \textbf{Team Page:}\enspace
  \href{https://home.polyadiclabs.com/}
       {\nolinkurl{home.polyadiclabs.com/}}
\end{center}

\end{abstract}

\section{Introduction}\label{sec:introduction}
Continuous diffusion language models (CDLMs) generate text by evolving continuous representations and decoding them into tokens.
Recent methods, including Plaid, LangFlow, RePlaid, and ELF, explore different choices of latent representation, training objective, and sampling procedure~\citep{gulrajani2023plaid,chen2026langflow,yang2026replaid,hu2026elf,luo2026designing}.
Common metrics include validation perplexity (PPL) on held-out text and, for generated samples, generative perplexity (Gen PPL) and within-sequence unigram entropy.
These evaluations capture different aspects of performance and can rank the same models differently. 
For example, one model may achieve lower validation PPL, while another produces samples with lower Gen PPL at comparable entropy. 
Even within a single evaluation setting, a better score can be difficult to interpret as evidence of progress.
This raises a broader question for CDLM research: \textit{what constitutes progress, and how can we measure it reliably across models and samplers?}

Likelihood-based evaluation asks how much probability a model assigns to
held-out text. For an autoregressive (AR) model, validation perplexity (PPL)
can be computed directly from the model's token conditionals. For a latent
CDLM, evaluating the likelihood of a discrete sequence generally requires
integrating over continuous latent variables. In practice, reported
validation PPL is often obtained by exponentiating a variational upper
bound on the per-token negative log-likelihood (NLL). The tightness of
this bound depends on how well the reference distribution used for
evaluation approximates the model's latent posterior given the observed
text. 
For example, at a fixed checkpoint, changing the reference noise
scale can lower the reported PPL if it brings the reference closer to that
posterior. 
A lower reported PPL can therefore reflect higher likelihood on held-out text, a tighter bound, or both.

Generated-sample-based evaluation asks how closely a sampler's output distribution matches that of real text.  
Gen PPL measures how predictable the generated text is under an external language model, and within-sequence unigram entropy measures how varied token usage is within each sample. 
A sampler that generates the same fluent, lexically varied passage on every draw can achieve both low Gen PPL and high within-sequence entropy, even under complete collapse across samples. 
We illustrate five failure modes using controlled collections of 512 texts of 1,024 tokens each, constructed from OpenWebText passages.
The real-text reference has a Gen PPL of 15.14 under GPT-2-large and a mean within-sequence unigram entropy of 5.443 nats.

\begin{figure}[!htbp]
  \centering
  \includegraphics[width=0.7\linewidth]{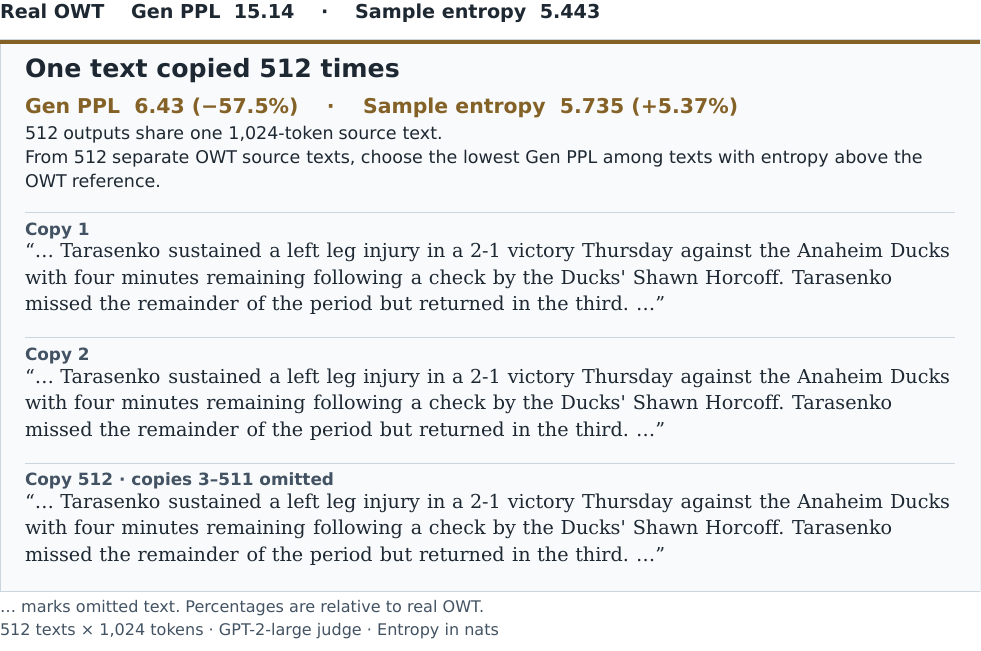}
  \caption{\textbf{Exact duplication across outputs.} From 512 candidate texts, we selected the passage with the lowest Gen PPL among those with above-reference entropy. Copies 1, 2, and 512 show the same excerpt.}
  \label{fig:intro-exact-duplication}
\end{figure}

Copying one selected passage 512 times gives a Gen PPL of 6.43 and an entropy of 5.735 nats, even though every output is identical (Figure~\ref{fig:intro-exact-duplication}).

\begin{figure}[!htbp]
  \centering
  \includegraphics[width=0.7\linewidth]{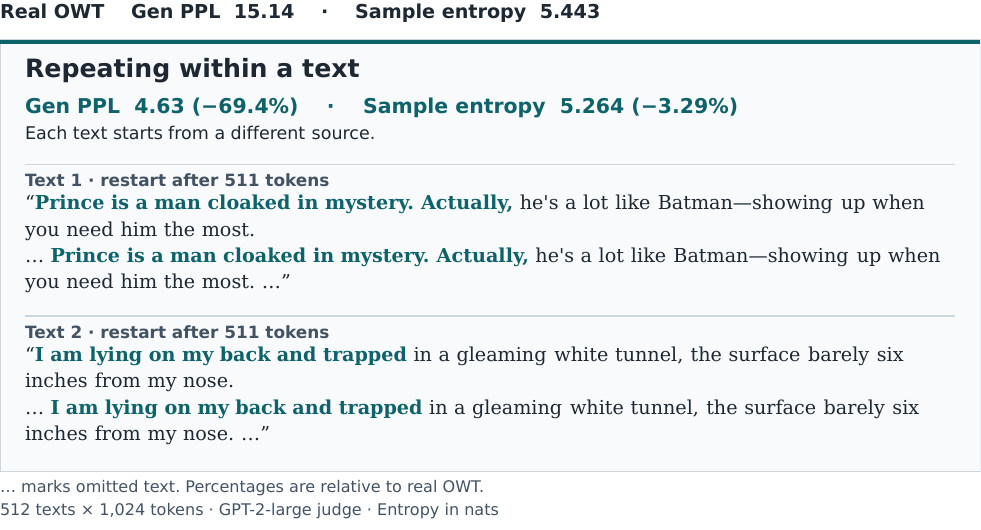}
  \caption{\textbf{Repetition within a text.} For each of two source texts, the excerpts show the opening and the point where it repeats.}
  \label{fig:intro-within-text-repetition}
\end{figure}

When each output repeats a passage from a different source, all 512 outputs are distinct. The judge can readily predict the repeated passage, bringing overall Gen PPL down to 4.63 (Figure~\ref{fig:intro-within-text-repetition}).

\begin{figure}[!htbp]
  \centering
  \includegraphics[width=0.7\linewidth]{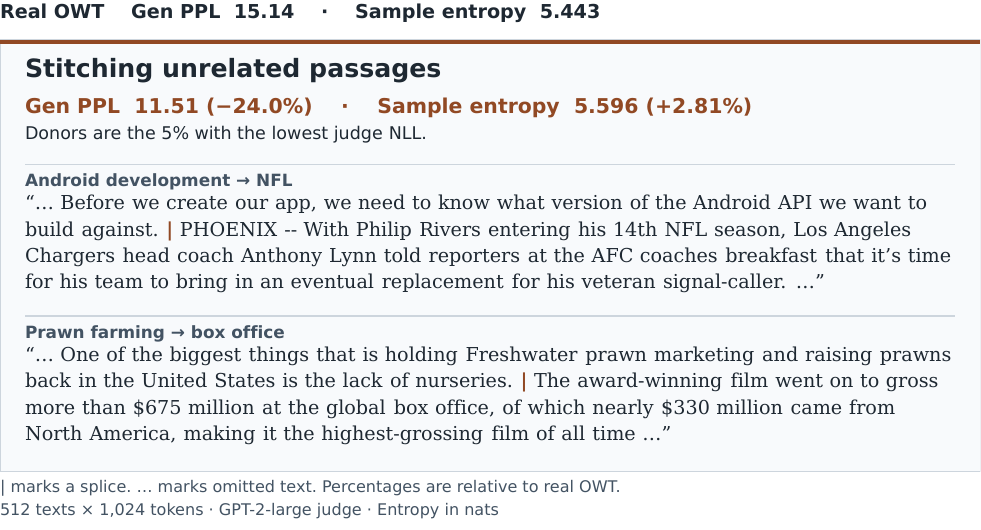}
  \caption{\textbf{Stitching unrelated passages.} Vertical bars mark joins between fragments of roughly 256 tokens. The donor texts come from the 5\% with the lowest judge NLL.}
  \label{fig:intro-unrelated-passages}
\end{figure}

Stitching together unrelated passages selected for low judge NLL yields a Gen PPL of 11.51 and an entropy of 5.596 nats, even when a text jumps from Android development to professional football (Figure~\ref{fig:intro-unrelated-passages}).

\begin{figure}[t]
  \centering
  \includegraphics[width=0.7\linewidth]{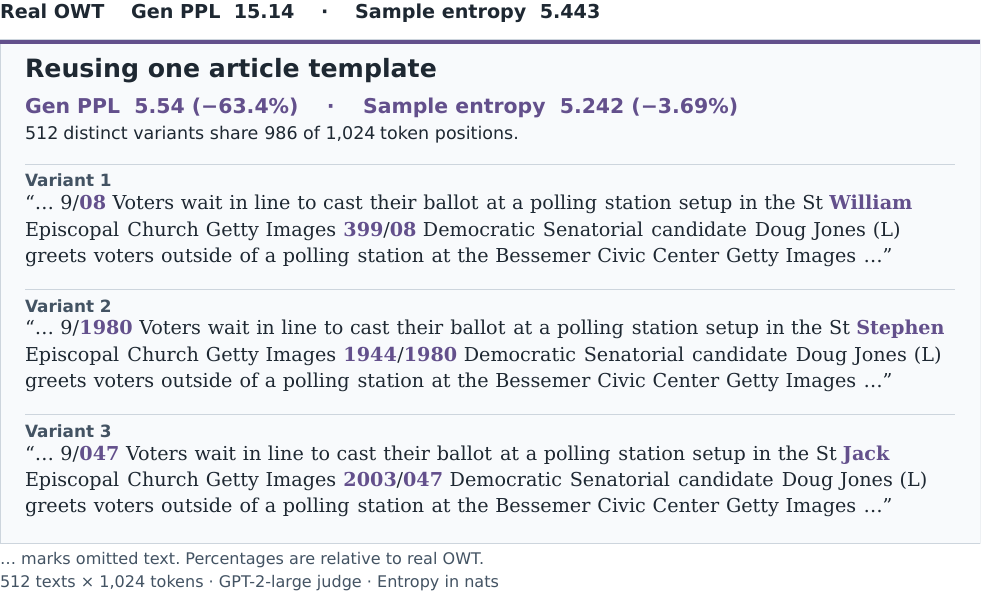}
  \caption{\textbf{Template collapse.} Bold text marks the names and numbers that change across three variants of the same article.}
  \label{fig:intro-template-collapse}
\end{figure}

Substituting names and numbers in a single article produces 512 distinct texts. The variants share 986 of their 1,024 token positions and all have the same unigram entropy (Figure~\ref{fig:intro-template-collapse}).

\begin{figure}[t]
  \centering
  \includegraphics[width=0.7\linewidth]{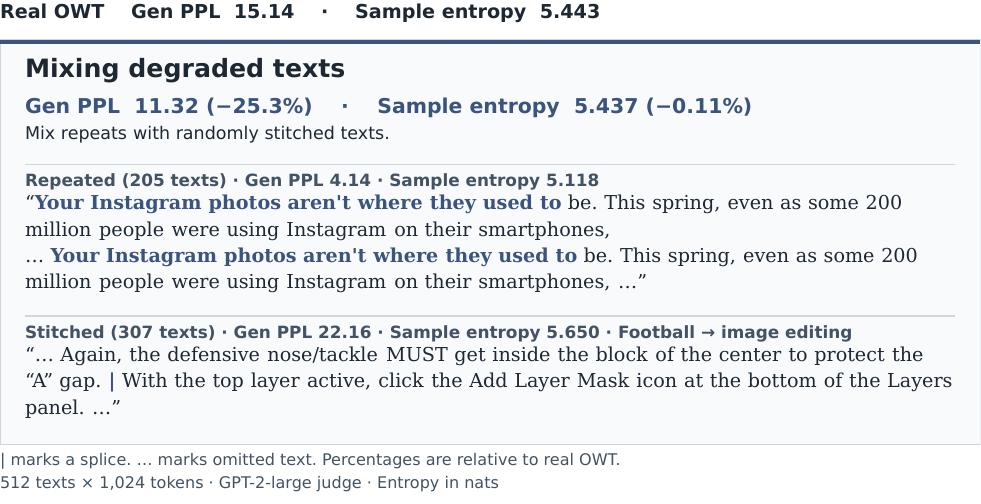}
  \caption{\textbf{Mixing two degradations.} The repetitive subset has lower mean unigram entropy than the reference, while the randomly stitched subset has higher mean entropy.}
  \label{fig:intro-mixture-degradations}
\end{figure}

A mixture of 205 internally repetitive texts and 307 texts stitched from random sources has a mean entropy of 5.437 nats, close to the real-text reference of 5.443, while Gen PPL falls by 25.3\% (Figure~\ref{fig:intro-mixture-degradations}).

A benchmark can guide CDLM development by giving researchers a common evaluation protocol and making clear what the resulting scores tell them about model fit and generated text.

We introduce \textbf{\bench}, a benchmark for evaluating CDLMs and their samplers, with an initial focus on unconditional generation on OpenWebText~\citep{gokaslan2019openwebtext}.
On held-out text, the benchmark specifies how to report and interpret perplexity bounds under matched scoring conditions.
For generation, it compares collections of generated and held-out texts using Kernel Text Discrepancy (KTD), with diagnostics for repetition, coverage, and validity.
Comparisons between sets of real texts provide a reference for ordinary sampling variation, and uncertainty estimates help interpret differences across models and
samplers. Our main contributions are:
\begin{enumerate}[leftmargin=*,itemsep=3pt]
  \item \textbf{Likelihood evaluation and comparison rules.}
  We characterize when reported perplexity bounds support conclusions about model fit and distinguish genuine tightening of the bound from apparent gains caused by stochastic log-weights. 
  An analytical counterexample and an archived LangFlow evaluation illustrate the limits of tuning the reference-noise scale as a way to assess the tightness of a bound.

  \item \textbf{A distributional evaluation protocol for generated text.}
  We characterize how Gen PPL and within-sequence entropy can miss collapse across samples. 
  In controlled OpenWebText experiments, MAUVE detects collapse, and its response to other perturbations depends on the text representation and the type of corruption.
  These findings inform the design of KTD, reference comparisons, and diagnostics in the benchmark.
\end{enumerate}

To support adoption of the benchmark, we will release evaluation code and open-source baseline reproductions through Open CDLM, beginning with LangFlow and RePlaid. These implementations will help researchers reproduce baseline results and evaluate new models and samplers under
the same protocol.

\section{Preliminary: Evaluation Objects and Scope}\label{sec:scope}
We distinguish three distributions. Let $P$ denote the target distribution of discrete texts. Let $P_\theta$ denote the discrete marginal of a specified generative model, including its continuous dynamics and text decoder. Finally, let $Q_{\theta,s}$ denote the output distribution of the implemented sampler with configuration $s$. Finite solver steps, temperature, self-conditioning, stopping rules, and final token selection can affect $Q_{\theta,s}$. Equality between $Q_{\theta,s}$ and $P_\theta$ requires justification; it is not implied by their use of the same checkpoint.

Likelihood evaluation measures how much probability $P_\theta$ assigns to held-out texts from $P$. Generation evaluation compares samples from $Q_{\theta,s}$ with a reference sample from $P$. A generation metric does not recover the unknown likelihood of the model, and a likelihood bound does not by itself characterize the behavior of every practical sampler.

For a latent model with continuous variable $z$ and normalized decoder $c_\theta$, the discrete probability is
\begin{equation}
 P_\theta(x)=\int p_\theta(z)c_\theta(x\mid z)\,\mathrm dz.
 \label{eq:discrete-marginal}
\end{equation}
The latent variable may represent an endpoint or an entire diffusion path. A continuous density evaluated at an embedding is not the probability of the associated text. The integration and decoder in \eqref{eq:discrete-marginal} are part of the model definition.

All logarithms are natural. For a fixed evaluation corpus $\mathcal D$, with scored token counts $L_x>0$, exact perplexity is
\begin{equation}
 \operatorname{PPL}_\theta(\mathcal D)
 =\exp\!\left(\frac{\sum_{x\in\mathcal D}-\log P_\theta(x)}
                    {\sum_{x\in\mathcal D}L_x}\right).
 \label{eq:corpus-ppl}
\end{equation}
We use total NLL divided by total scored tokens rather than an arithmetic mean of per-text perplexities. Comparisons require the same corpus, tokenization, context boundaries, and scoring convention. The initial benchmark scope is unconditional generation on OpenWebText (OWT), with LM1B as an extension. The controlled metric pilot uses 128-token source windows; the archived likelihood experiment uses 1024-token rows. These experiments address different questions and are not combined into one model-performance table.

\section{What CDLM Perplexity Can Identify}\label{sec:likelihood}
\subsection{Discrete likelihood, continuous density, and variational slack}
Fix the generative model in \eqref{eq:discrete-marginal} and a text with $P_\theta(x)>0$. For a normalized reference $r(z\mid x)$, define the negative evidence lower bound (NELBO)
\begin{equation}
 U_r(x)=\E_{r(z\mid x)}
 \big[\log r(z\mid x)-\log p_\theta(z)-\log c_\theta(x\mid z)\big].
 \label{eq:negative-elbo}
\end{equation}
For this fixed $x$, normalizing the joint density over $z$ gives the model posterior
\[
 p_\theta(z\mid x)
 =\frac{p_\theta(z)c_\theta(x\mid z)}{\int p_\theta(z')c_\theta(x\mid z')\,\mathrm dz'}
 =\frac{p_\theta(z)c_\theta(x\mid z)}{P_\theta(x)}.
\]
Assuming that the expectations exist and $r$ is absolutely continuous with respect to the posterior, Bayes' rule gives
\begin{equation}
 \boxed{U_r(x)=-\log P_\theta(x)+G_r(x),\qquad
 G_r(x)=\KL\!\left(r(z\mid x)\Vert p_\theta(z\mid x)\right)\geq0.}
 \label{eq:gap}
\end{equation}
The gap vanishes precisely when the reference equals the model posterior almost everywhere. A known Gaussian corruption distribution is not automatically this posterior, and token probabilities output by a denoiser are not the continuous posterior conditioned on the complete observed text.

Equation~\eqref{eq:gap} is a standard variational identity, not a new bound. Its role here is to identify the objects that a CDLM benchmark must distinguish. Endpoint bounds for flow models and path-space bounds for variational diffusion models instantiate it with different latent variables \citep{chen2026langflow,kingma2021vdm,yang2026replaid}. The gap also differs from a missing conversion constant between continuous and discrete likelihood. The decoder and reference-density terms already belong to the valid bound; dropping them changes the quantity being evaluated. Appendix~\ref{app:density} makes this distinction explicit through a joint KL decomposition.

Exponentiating the corpus bound yields
\begin{equation}
 \frac{\operatorname{PPL}_{\mathrm{bound},r}}
      {\operatorname{PPL}_{\theta}}
 =\exp\!\left(\frac{\sum_{x\in\mathcal D}G_r(x)}
                   {\sum_{x\in\mathcal D}L_x}\right).
 \label{eq:ppl-gap}
\end{equation}
Thus $U_r$ is a sequence-level NLL bound; bound perplexity is its token-normalized exponential. A gap in nats per token becomes a multiplicative gap in perplexity. Without additional inference or assumptions, the reported bound alone does not identify this multiplier. This is an identifiability limitation, not a statement that the gap can never be estimated by stronger methods.

\subsection{Importance weighting and multi-sample upper bounds}\label{sec:multisample}
Importance weighting expresses the discrete marginal as an expectation under a tractable reference. In addition to the conditions for a finite NELBO, assume that $r(\cdot\mid x)$ covers the support of the model posterior. Then
\begin{align}
 w_r(z,x)&=\frac{p_\theta(z)c_\theta(x\mid z)}{r(z\mid x)},
 &\E_{Z\sim r(\cdot\mid x)}w_r(Z,x)&=P_\theta(x).
 \label{eq:importance-identity}
\end{align}
The denominator corrects for sampling from $r$ rather than integrating with respect to the latent base measure. Consequently, the same marginal probability is recovered for any such reference, whereas $U_r=\E_r[-\log w_r]$ can vary with the reference because the logarithm is inside the expectation.

For $N$ independent latent draws $Z_{1:N}\sim r(\cdot\mid x)^{\otimes N}$, define
\begin{align}
 U_{r,N}(x)
 &=-\E_{r^{\otimes N}}\log\left(\frac1N\sum_{n=1}^Nw_r(Z_n,x)\right) \notag\\
 &=-\E_{r^{\otimes N}}\log\left(\frac1N\sum_{n=1}^N
 c_\theta(x\mid Z_n)\exp\!\big(\Lambda_r(Z_n,x)\big)\right),
 \label{eq:multisample-bound}\\
 \Lambda_r(z,x)&=\log p_\theta(z)-\log r(z\mid x).
 \label{eq:pointwise-density-ratio}
\end{align}
In an endpoint formulation, one may take $z=x_\tau$ and $r(z\mid x)$ to be the forward corruption density at a fixed $\tau$. The ratio $\Lambda_r$ must be evaluated at each sampled endpoint. It is a pointwise log-density ratio, distinct from a change in the expected bound. Averaging over target texts gives the population inequality
\begin{align}
 \E_{X\sim P}[-\log P_\theta(X)]
 &=-\E_{X\sim P}\log\E_{Z\sim r(\cdot\mid X)}w_r(Z,X) \notag\\
 &\leq-\E_{X\sim P}\E_{Z_{1:N}\sim r(\cdot\mid X)^{\otimes N}}
 \log\left(\frac1N\sum_{n=1}^Nw_r(Z_n,X)\right) \notag\\
 &=\E_{X\sim P}U_{r,N}(X).
 \label{eq:population-multisample}
\end{align}
The expectation over the latent draws on the right is essential: an individual realization of the sample average need not upper-bound NLL after taking its negative logarithm.

\begin{proposition}[Importance-weighted tightening]\label{prop:multisample}
For exact weights with the stated support conditions and finite expectations,
\begin{equation}
 -\log P_\theta(x)\leq U_{r,N+1}(x)\leq U_{r,N}(x)\leq U_{r,1}(x)=U_r(x).
 \label{eq:multisample-order}
\end{equation}
If the logarithms of the sample-mean weights are uniformly integrable, $U_{r,N}(x)\to-\log P_\theta(x)$. One sufficient condition is $0<a\leq w_r(Z,x)\leq b<\infty$ almost surely for constants depending on $x$.
\end{proposition}
These are the standard importance-weighted bounds \citep{burda2016iwae}. Appendix~\ref{app:multisample-proof} gives the proof and the assumptions needed to pass from almost-sure convergence of the weights to convergence of the expected logarithm.

Define the remaining gap and the improvement over the single-sample bound by
\begin{align}
 G_{r,N}(x)&=U_{r,N}(x)+\log P_\theta(x)\geq0,
 &G_{r,1}(x)&=G_r(x), \label{eq:multisample-gap}\\
 I_{r,N}(x)&=U_r(x)-U_{r,N}(x)
 =G_r(x)-G_{r,N}(x)\in[0,G_r(x)].
 \label{eq:multisample-improvement}
\end{align}
Increasing $N$ tightens the evaluation objective for the same model; it does not change that model's true NLL. A positive ideal $I_{r,N}$ lower-bounds the original gap without requiring the unknown NLL, because that NLL cancels in the difference. For $N>1$, $G_{r,N}$ is not the original single-latent posterior KL; Appendix~\ref{app:multisample-proof} provides its joint-space KL representation. For a corpus,
\begin{equation}
 \frac{\operatorname{PPL}_{\mathrm{bound},r,N}(\mathcal D)}
 {\operatorname{PPL}_{\theta}(\mathcal D)}
 =\exp\!\left(\frac{\sum_{x\in\mathcal D}G_{r,N}(x)}
 {\sum_{x\in\mathcal D}L_x}\right),\qquad
 \operatorname{PPL}_{\mathrm{bound},r,N}
 =\exp\!\left(\frac{\sum_x U_{r,N}(x)}{\sum_x L_x}\right).
 \label{eq:multisample-ppl}
\end{equation}
The sequence weights must be combined before token normalization. Appendix~\ref{app:weight-aggregation} derives the corresponding estimator and explains why independently combining token weights changes the target.

\subsection{Comparison with another CDLM or an autoregressive model}
Let $\mathcal U_A=\mathcal L_A+\mathcal G_A$ denote a token-normalized bound, exact NLL, and gap for model $A$, with analogous notation for $B$. Then
\begin{equation}
 \mathcal U_A-\mathcal U_B
 =(\mathcal L_A-\mathcal L_B)+(\mathcal G_A-\mathcal G_B).
 \label{eq:ranking}
\end{equation}
Matching the evaluation code does not guarantee matching posterior mismatch across checkpoints. Consequently, a change in the bound does not isolate a change in exact NLL. The same decomposition holds for multi-sample bounds with their respective $G_{r,N}$: using a common $N$ does not make the remaining gaps equal.

\begin{proposition}[A bound ranking does not identify a likelihood ranking]
\label{prop:ranking}
There exist normalized latent-variable models with the same reported values $\mathcal U_A<\mathcal U_B$ for which $\mathcal L_A<\mathcal L_B$, and other normalized models with those same reported values for which $\mathcal L_A>\mathcal L_B$.
\end{proposition}
Appendix~\ref{app:ranking-proof} gives an explicit construction with Gaussian posteriors and a common reference. The proposition concerns what the reported numbers logically determine; it is not evidence that a particular published model ranking is reversed.

Comparison with exact AR NLL has a useful asymmetry. Under a matched scoring protocol, a valid ideal bound satisfying
\begin{equation}
 \mathcal U_{\mathrm{CDLM}}<\mathcal L_{\mathrm{AR}}
 \quad\Longrightarrow\quad
 \mathcal L_{\mathrm{CDLM}}<\mathcal L_{\mathrm{AR}}
 \label{eq:ar-direction}
\end{equation}
certifies an improvement in the underlying CDLM likelihood. If the CDLM bound is larger, the reverse conclusion does not follow: the unobserved gap may account for the difference. Bound perplexity therefore remains informative, while it cannot be treated as an exact perplexity for unrestricted two-way ranking. Finite estimates require an additional assessment of estimation error before applying \eqref{eq:ar-direction}.

\subsection{Computing a bound introduces separate errors}
A reported single-sample estimate can be organized as
\begin{equation}
 \widehat{\mathcal U}
 =\mathcal L_\theta+\mathcal G_r
   +b_{\mathrm{impl}}+b_{\mathrm{num}}+\eta_{\mathrm{MC}}.
 \label{eq:error-decomposition}
\end{equation}
Here $b_{\mathrm{impl}}$ is the discrepancy between the specified ideal objective and the objective implemented in exact arithmetic; $b_{\mathrm{num}}$ is the additional bias of the numerical procedure; and $\eta_{\mathrm{MC}}$ is fluctuation around that procedure's expectation. This decomposition presumes that the intended target and the intervening procedures have been specified. Unlike $\mathcal G_r$, the implementation and numerical terms need not be nonnegative. A single estimate is therefore not guaranteed to upper-bound exact NLL.

For an ODE density estimator, the audit must account for both state integration and the divergence integral \citep{chen2018neuralode,grathwohl2019ffjord}. It must also specify which computation is differentiated when the network uses self-conditioning. For a path-space diffusion objective, the corresponding audit concerns the transition or continuous-time objective, endpoint terms, and time/noise sampling. These are different estimation procedures. In particular, ODE and reverse-SDE models induced by an approximate network need not have identical marginals; the ideal equivalence relies on the appropriate score and boundary conditions \citep{song2021sde}. A difference between their reported bounds cannot automatically be attributed to variational slack for one shared model.

The benchmark consequently records the target joint model, decoder, bound type, estimator, precision, solver or time-sampling settings, and random seeds alongside every likelihood result. We compare and quantify uncertainty in nats per token before exponentiating. Even an unbiased NLL estimate becomes a biased perplexity estimate after exponentiation in general.

\subsection{Stochastic log-weights can imitate bound tightening}\label{sec:noisy-weights}
The distinction between objective and estimator becomes especially important for $N>1$. Let an estimated log-weight satisfy
\begin{equation}
 \widehat\ell_r(z,x)=\log w_r(z,x)+\xi,\qquad
 \E[\xi\mid z,x]=0.
 \label{eq:noisy-log-weight}
\end{equation}
Even under this favorable zero-bias assumption, Jensen's inequality gives
\begin{equation}
 \E[\exp(\widehat\ell_r)\mid z,x]
 =w_r(z,x)\E[e^\xi\mid z,x]\geq w_r(z,x).
 \label{eq:noisy-weight-bias}
\end{equation}
Thus an unbiased log-density estimate does not supply an unbiased importance weight. A Hutchinson trace estimator can be unbiased along a fixed trajectory while its exponentiated density estimate has this problem. Numerical integration and finite precision can add further bias.

\begin{proposition}[Apparent tightening with zero variational gap]\label{prop:noisy-tightening}
Suppose $r(z\mid x)=p_\theta(z\mid x)$ and the additional log-weight errors are independent $\xi_n\sim\mathcal N(0,v)$. All exact bounds equal $-\log P_\theta(x)$, but the noisy estimator satisfies
\begin{equation}
 -\log\left(\frac1N\sum_{n=1}^N e^{\widehat\ell_{r,n}}\right)
 \xrightarrow[N\to\infty]{\mathrm{a.s.}}
 -\log P_\theta(x)-\frac v2.
 \label{eq:noisy-zero-gap}
\end{equation}
Its expectation is nonincreasing in $N$, starting at the exact NLL for $N=1$.
\end{proposition}
Appendix~\ref{app:noisy-weights} proves the claim and separates latent-draw uncertainty from conditional density-estimation noise. The example establishes a failure mechanism, not the noise distribution or bias magnitude of a particular checkpoint. Its per-token limiting offset is $v/(2L_x)$; estimating a variance alone does not justify a variance-based correction to a practical result.

In an ODE calculation, $N$ denotes latent draws, $M$ denotes trace probes per density estimate, and $S$ denotes integration steps. They control different approximations. An $N$ curve should be accompanied by independent repeated density estimates at fixed latent states, paired solver and precision checks, and diagnostics of normalized weights, such as
\begin{equation}
 \widetilde w_n=\frac{w_n}{\sum_jw_j},\qquad
 \operatorname{ESS}=\frac{1}{\sum_n\widetilde w_n^2},\qquad
 w_{\max}=\max_n\widetilde w_n.
 \label{eq:importance-ess}
\end{equation}
Low ESS indicates concentration among the sampled weights; it does not invalidate the exact bound or measure its remaining gap. When the weights are estimated, these diagnostics also depend on density noise. Moreover, pooling the same positive weights into larger groups can produce a decreasing curve by concavity alone, regardless of whether the weights are correct (Appendix~\ref{app:weight-aggregation}). A decreasing empirical curve therefore needs numerical and sampling controls before it can be interpreted through \eqref{eq:multisample-improvement}.

\subsection{What reference-family tightening does and does not show}
One can seek evidence of a gap without knowing the exact likelihood. Fix the model and compare a default reference $r_1$ with a candidate $r_\eta$. Equation~\eqref{eq:gap} implies
\begin{equation}
 \Delta_\eta(x)=U_{r_1}(x)-U_{r_\eta}(x)
 =G_{r_1}(x)-G_{r_\eta}(x)\leq G_{r_1}(x).
 \label{eq:recoverable-gap}
\end{equation}
\begin{proposition}[Recoverable gap]\label{prop:recoverable}
A positive ideal average improvement obtained by changing only the reference lower-bounds the default average gap. Failure to find an improvement does not certify a small gap.
\end{proposition}
\begin{proof}
The first claim follows by averaging \eqref{eq:recoverable-gap}, since the candidate gap is nonnegative. The second follows because two references can have equally large posterior mismatch, including when every tested candidate is worse than the default.
\end{proof}

Candidate selection and confirmation must use separate data. If a lower confidence limit for the measured improvement is $c$ and the absolute bias of the difference is bounded by $\varepsilon$, one can support a gap of at least $c-\varepsilon$ at the corresponding confidence level. Agreement between two numerical settings alone does not establish such a bias bound. The experiment in Section~\ref{sec:ppl-results} is therefore reported as a reference-scale diagnostic, without treating its sampling interval as a certificate of numerical accuracy.

Reference adaptation and multi-sample evaluation are complementary: the former changes $r$, while the latter changes $N$ for a fixed $r$. Both retain $P_\theta$ when implemented with the correct density ratio. Appendix~\ref{app:reference-density} derives normalized mean and covariance adaptations and explains why replacing a Gaussian sample's squared distance by its expectation is valid for the single-sample expected NELBO but not for constructing multi-sample weights. A separate upper/lower-bound identity on texts drawn from the model itself appears in Appendix~\ref{app:model-sandwich}; that diagnostic concerns a different text distribution from held-out evaluation.

\subsection{An optimal noise scale does not identify the total gap}
Consider a model posterior and a reference family
\begin{equation}
 p_\theta(z\mid x)=\mathcal N(\mu+\delta,\sigma^2 I_D),
 \qquad r_\rho(z\mid x)=\mathcal N(\mu,\rho^2\sigma^2 I_D).
\end{equation}
Their KL divergence is
\begin{equation}
 G_\rho(x)=\frac{\|\delta\|^2}{2\sigma^2}
          +\frac D2(\rho^2-1-2\log\rho).
 \label{eq:gaussian-gap}
\end{equation}
\begin{proposition}[Scale diagnostics leave an unidentified offset]
\label{prop:scale}
For every mean mismatch $\delta$, the minimizing scale in \eqref{eq:gaussian-gap} is $\rho=1$. The entire relative curve $G_\rho-G_1$ is independent of $\delta$, while $G_1$ can be arbitrarily large.
\end{proposition}
This follows directly from \eqref{eq:gaussian-gap}; the derivative of its scale term is $D(\rho-\rho^{-1})$. When $D=Ld$, a local expansion gives
\begin{equation}
 \frac{G_{1+t}-G_1}{L}
 =d\left(t^2-\frac{t^3}{3}+\frac{t^4}{4}+O(t^5)\right).
 \label{eq:scale-expansion}
\end{equation}
A quadratic coefficient near $d$ is thus compatible with zero gap or with substantial unobserved mean mismatch. The coefficient is $d$, whereas the second derivative at the optimum is $2d$. A general covariance introduces further unidentified terms, as detailed in Appendix~\ref{app:scale-proof}.

\begin{figure}[tb]
 \centering
 \includegraphics[width=\linewidth]{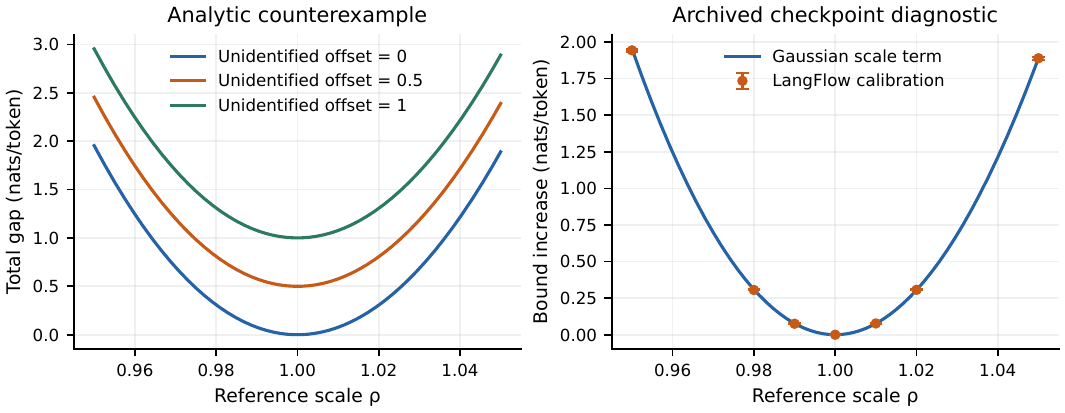}
 \caption{\textbf{A scale curve cannot recover its missing offset.} Left: analytic per-token gaps with $d=768$ and three chosen mean-mismatch offsets. These are constructed examples, not checkpoint measurements. Right: the archived LangFlow calibration experiment, showing the change relative to $\rho=1$ and the analytic scale term. Error bars are paired row-bootstrap intervals. Similar relative curvature does not identify the total posterior gap.}
 \label{fig:likelihood}
\end{figure}

These results lead to a concrete interpretation rule: report likelihood bounds as bounds on specified models, and report reference-family improvements as evidence about recoverable slack. Neither a tight-looking scale curve nor a favorable point estimate justifies replacing the bound by an asserted exact PPL.

\section{Evaluating the Generated Distribution}\label{sec:generation}
\subsection{Why generative perplexity and sample entropy are insufficient}
Let $J$ be a fixed external autoregressive judge, and let $\ell_J(x)$ be its summed NLL over the scored tokens of text $x$. For outputs $Y=\{y_1,\ldots,y_m\}$,
\begin{equation}
 \GenPPL(Y)=\exp\!\left(
 \frac{\sum_{j=1}^m\ell_J(y_j)}{\sum_{j=1}^m T_J(y_j)}
 \right),
 \qquad
 \Hrow(Y)=\frac1m\sum_{j=1}^m H(\widehat p_{y_j}),
 \label{eq:legacy}
\end{equation}
where $T_J(y)$ is the judge's scored token count and $\widehat p_y$ is the text's empirical unigram distribution under the stated entropy tokenizer. The sample-entropy definition in LangFlow is of this within-sequence form \citep{chen2026langflow}. Gen PPL measures how predictable outputs are to $J$; it is not the generating model's held-out perplexity.

For $Y=\{y^\star,\ldots,y^\star\}$, both quantities in \eqref{eq:legacy} equal their value on $y^\star$. The corresponding point-mass generator has sequence entropy zero. If $y^\star$ has low judge NLL and a varied token histogram, both conventional scores can look favorable even though the model produces no variation across outputs. In Section~\ref{sec:stress-results}, a selected real text realizes this score-level failure empirically. Its selection does not constitute a human quality assessment.

The limitation is broader than a specially chosen single text.
\begin{proposition}[Two aggregate scores permit small support]\label{prop:small-support}
For any distribution on a finite text collection with a common positive judge-scored length, there is a distribution supported on at most three of those texts with exactly the same population Gen PPL and mean within-sequence unigram entropy.
\end{proposition}
The result follows by matching the two-dimensional mean of $(\ell_J(x)/T_J,H(\widehat p_x))$; an elementary proof appears in Appendix~\ref{app:moment-proof}. The matching distribution has sequence entropy at most $\log 3$, regardless of the size of the original collection. Increasing the number of evaluation samples estimates these two moments more precisely but does not restore the distributional information they omit. The fixed-length assumption is material to this two-moment formulation.

\subsection{Kernel Text Discrepancy}\label{sec:ktd}
Kernel Text Discrepancy (KTD) compares a corpus of generated texts with a corpus of real texts. We represent each text by the distribution of its contextual token states and its final token state, then measure the squared maximum mean discrepancy (MMD) between the two distributions of text representations \citep{gretton2012kernel}.

\paragraph{Representing a text.}
Let $h_1(x),\ldots,h_T(x)\in\mathbb R^{1280}$ be the final-layer states of a frozen GPT-2-large encoder for the $T$ retained tokens of $x$, excluding padding. We draw $R=128$ independent standard Gaussian vectors, normalize each to unit length, and reuse the resulting directions $v_1,\ldots,v_R$ for every text (seed 17). For direction $j$, let $\widehat F_{xj}$ be the empirical CDF of the $T$ projections $v_j^\top h_t(x)$. We summarize this distribution in $B=64$ equal-probability bins. Each entry is the average of the empirical quantile function over its bin,
\begin{equation}
 q_{jb}(x)=B\int_{(b-1)/B}^{b/B}\widehat F_{xj}^{-1}(u)\,\mathrm du,
 \qquad j=1,\ldots,R,\quad b=1,\ldots,B.
 \label{eq:ktd-quantiles}
\end{equation}
This definition also applies when $T<B$. Flattening these entries gives $q(x)\in\mathbb R^{8192}$, which records variation among tokens within the text. The normalized last-token state $\ell(x)=h_T(x)/\|h_T(x)\|_2$ encodes the retained sequence through the encoder's causal attention.

\paragraph{Calibrating the kernel.}
On a separate corpus of real texts $C=\{c_i\}$, define $s(f)=\operatorname{median}_{i<j}\|f(c_i)-f(c_j)\|_2$, using the lower middle distance when the number of pairs is even. We scale the two feature blocks separately and concatenate them,
\begin{equation}
 \phi(x)=\left[\frac{\sqrt{3/4}}{s(q)}\,q(x),\
               \frac{\sqrt{1/4}}{s(\ell)}\,\ell(x)\right],
 \qquad \sigma=\tfrac12 s(\phi).
 \label{eq:ktd-features}
\end{equation}
The coefficients $3/4$ and $1/4$ weight the two blocks' contributions to the squared distance. We fix the preprocessing, directions, scales, and bandwidth across all systems in a comparison and use the Gaussian kernel
\begin{equation}
 k_\sigma(x,y)=\exp\!\left(-\frac{\|\phi(x)-\phi(y)\|_2^2}{2\sigma^2}\right).
 \label{eq:feature-kernel}
\end{equation}

\paragraph{Scoring two corpora.}
Let $X,X'$ be independent draws from the real-text distribution $P$ and $Y,Y'$ independent draws from the generator $Q=Q_{\theta,s}$, independent of $X,X'$. The population quantity is
\begin{equation}
 D_{k_\sigma}^2(P,Q)=
 \E k_\sigma(X,X')+\E k_\sigma(Y,Y')-2\E k_\sigma(X,Y).
 \label{eq:mmd-pop}
\end{equation}
For reference texts $\mathcal X=(x_1,\ldots,x_n)$ and generated texts $\mathcal Y=(y_1,\ldots,y_m)$ with $n,m\geq2$, we report the U-statistic on this squared scale,
\begin{align}
 \operatorname{KTD}(\mathcal X,\mathcal Y)=\widehat D_u^2
 &=\frac{\sum_{i\ne i'}k_\sigma(x_i,x_{i'})}{n(n-1)}
   +\frac{\sum_{j\ne j'}k_\sigma(y_j,y_{j'})}{m(m-1)} \notag\\
 &\quad-\frac{2}{nm}\sum_{i,j}k_\sigma(x_i,y_j).
 \label{eq:mmd-u}
\end{align}
Each within-corpus sum excludes self-pairs, and the cross-corpus sum includes every pair. Conditional on the independently calibrated kernel, the estimate is unbiased for independent, identically distributed draws in each corpus. Sampling variation can give a negative estimate, which we retain.

A smaller population KTD indicates closer agreement in the retained representation. For example, an encoder that truncates texts to a fixed prefix cannot distinguish edits beyond that prefix. Corpus agreement and human preference can also diverge if a generator avoids poor material present in the reference. These considerations motivate the complementary diagnostics below.

\subsection{Diagnostics for collapse, coverage, and local failures}
We accompany KTD with measurements that make important failure modes directly observable. Exact and near-duplicate rates characterize concentration on a small output library. Within-text $4$-gram repetition detects local loops. Invalid outputs, effective lengths, and special-token behavior are reported explicitly; silently discarding empty or malformed outputs would change the evaluated generator.

We also report
\begin{equation}
 \JSDrows(Y)=H\!\left(\sum_j w_j\widehat p_{y_j}\right)
                 -\sum_j w_jH(\widehat p_{y_j}),
 \qquad w_j=\frac{|y_j|}{\sum_i|y_i|}.
 \label{eq:jsd}
\end{equation}
For equal lengths this is pooled unigram entropy minus $\Hrow$. It vanishes when all outputs have the same token histogram, including exact replication. It can also vanish for differently ordered texts, so it is not sequence entropy and is not sufficient as a distribution metric. Feature-space precision, recall, density, and coverage provide complementary neighborhood summaries \citep{naeem2020reliable}, with neighborhood size fixed and reported.

Rare failures require explicit event or subgroup measurements. For the mixture $Q_\epsilon=(1-\epsilon)P+\epsilon R$, linearity of kernel means gives
\begin{equation}
 D_k^2(P,Q_\epsilon)=\epsilon^2D_k^2(P,R).
 \label{eq:contamination}
\end{equation}
For an RBF kernel, this is at most $2\epsilon^2$. A small overall discrepancy can therefore coexist with a consequential rare error. Likewise, a generator can completely miss a low-probability mode while having small MMD. Appendix~\ref{app:mmd-limits} gives the corresponding example with infinite forward KL.

The protocol consequently reports predeclared attributes and subgroup frequencies, together with within-subgroup discrepancy when enough samples exist. A missing subgroup is recorded as missing coverage rather than omitted from analysis. Acceptable changes in duplication, validity, or important subgroup frequencies must be set from task requirements and independent calibration. A score improvement is accompanied by these changes instead of being converted into an unconditional overall winner.

\subsection{Statistical protocol for research iteration}
Each comparison fixes the target split, number of outputs, length convention, judge, representation, and kernel. A second held-out real sample of equal size is evaluated against the scoring reference. This real-versus-real result is a finite-sample control, not a deterministic lower bound on the numerical score.

For independent two-sample testing with a fixed kernel, a permutation test can assess equality of feature distributions. If $B$ random permutations are used, we report the Monte Carlo value
\begin{equation}
 \widehat p=\frac{1+\sum_{b=1}^B
     \mathbf 1\{\widehat D_{u,b}^2\geq\widehat D_{u,\mathrm{obs}}^2\}}
     {B+1}.
 \label{eq:permutation}
\end{equation}
The smallest attainable value is $1/(B+1)$. The permutation distribution describes variation under the null. Estimating uncertainty in a difference between two model scores requires a sampling design for that comparison.

Model comparisons use common reference samples and report score differences across independent generation runs or appropriately paired analyses. Resampling respects the actual independent unit, preferably documents when document identifiers are available. Near the null, the degeneracy of the MMD U-statistic requires suitable calibration rather than a generic Gaussian interpretation of a few null standard deviations.

Finally, exported samples are not treated as new independent observations merely because a file is copied. Natural duplicate outputs from independent generation remain in the evaluation, since deduplication would alter $Q_{\theta,s}$. Appendix~\ref{app:replication} shows why deterministic duplication of a sample file changes the U-statistic's finite-sample correction even though its empirical probability distribution is unchanged.

\section{The Open CDLM Library and Benchmark Design}\label{sec:library}
The purpose of the library is to make the evaluation object reproducible: a model specification, a checkpoint, a sampler configuration, and a measurement protocol. A common interface is useful only if it preserves the distinctions identified above. In particular, a sampler setting must not silently redefine a likelihood model, and a preprocessing change must not appear as an unexplained metric improvement.

\subsection{Model, sampling, and evaluation components}
The design separates four components. A data component specifies source splits, tokenizer revisions, document boundaries, packing, and evaluation lengths. A model component contains the embedding, network, training objective, noise schedule, self-conditioning rule, and decoder. A sampling component records the solver, step schedule, number of network evaluations, temperature, stopping behavior, and token selection. An evaluation component consumes saved outputs or a declared likelihood interface and produces measurements with their metadata.

The evaluation package provides a shared sample interface, cached judge features, Gen PPL, entropy decomposition, KTD, neighborhood diagnostics, repetition measurements, and MAUVE. Its likelihood modules and the archived scale study provide a starting point for likelihood auditing. The benchmark specification strengthens this core by requiring independent kernel calibration, explicit uncertainty semantics, and versioned output manifests.

The model component is designed to support unofficial implementations of LangFlow and RePlaid, including training, sampling, and evaluation configurations. Upstream methods and available implementations are credited \citep{chen2026langflow,yang2026replaid,langflowrepo2026}. The common interface separates model-specific computation from data handling, sampling orchestration, and evaluation. Loading an existing checkpoint into an adapter validates compatibility only; it is not evidence of reproducing the original training procedure.

\subsection{Original-recipe reproduction and standardized comparison}
The benchmark defines two separate tracks. In the \emph{reproduction track}, each model follows its original recipe as closely as the available specification permits. The report records any departure in data preparation, architecture, objective, optimization, or evaluation. It compares training curves, reported evaluation quantities, and sampling behavior with the corresponding source results. Reproduction fidelity is assessed before interpreting downstream metric differences.

In the \emph{standardized comparison track}, models share the evaluation corpus, sample count, length policy, preprocessing, judge, and measurement protocol. Training compute, model size, data exposure, and inference cost are either matched or reported explicitly. Sampler experiments hold the checkpoint fixed; model experiments identify the differing checkpoints and training budgets. This track measures performance under the benchmark's common conditions. It is not presented as a reproduction of every original paper's number.

OWT is the primary dataset, followed by an LM1B extension. AR and discrete diffusion controls provide reference points under the same generation protocol. A quality--cost comparison records wall-clock latency, throughput, hardware, precision, peak memory, and network evaluations. Step counts alone are insufficient because self-conditioning and different solver stages can require different amounts of computation.

\subsection{Reproducible artifacts}
Each benchmark result is designed to link to a configuration, source version, checkpoint hash, data identity, sampling seeds, output hash, and table-generation command. Saved sample records permit alternative metrics to be evaluated on identical outputs. A separate likelihood record identifies the target joint model and its estimator. A reproducibility bundle consists of these records, training logs, and checkpoint acquisition instructions.

\section{Experiments}\label{sec:experiments}
The experiments address two questions: what a reference-scale search reveals about an NLL bound, and which generation failures are visible to the proposed measurements. Results use archived likelihood measurements and controlled text constructions, with all tables and figures generated from the corresponding experimental records.

\subsection{A reference-scale diagnostic on LangFlow}\label{sec:ppl-results}
\paragraph{Setup.}
The archived experiment freezes a LangFlow OWT checkpoint, its decoder, continuous dynamics, and endpoints. It changes only the standard deviation of the forward endpoint reference,
\begin{equation}
 r_\rho(z_0\mid x)=
 \mathcal N\!\left(\alpha_0 e(x),\rho^2\sigma_0^2 I_D\right).
 \label{eq:scale-family}
\end{equation}
The reference log-density is changed consistently, including the $-D\log\rho$ term. This intervention changes the inference reference rather than the generator's temperature. The rows contain $L=1024$ GPT-2 tokens, the embedding dimension is $d=768$, and $D=786432$. Density evaluation uses 128 Heun steps with one Gaussian Hutchinson probe per function evaluation. Appendix~\ref{app:likelihood-protocol} records the estimator and numerical controls.

The calibration split contains 64 rows with four draws per row and scale. It evaluates $\rho\in\{0.95,0.98,0.99,1,1.01,1.02,1.05\}$. The disjoint evaluation split contains 256 rows with six draws per condition. Shared random draws pair each candidate with the baseline. Reported intervals use 20,000 bootstrap resamples of row-level paired means. They reflect sampling variation and do not include unknown systematic numerical bias or possible dependence between rows from the same document.

\paragraph{Calibration and held-out evaluation.}
The default scale $\rho=1$ is the best of the seven calibration points. The local quadratic coefficient is $766.13$, close to the analytic coefficient $d=768$ in \eqref{eq:scale-expansion}. Figure~\ref{fig:likelihood} shows the relative curve. This resemblance does not identify the total gap, as Proposition~\ref{prop:scale} establishes.

\begin{table}[!htbp]
 \centering\small
 \caption{\textbf{Held-out reference-scale evaluation.} $\Delta_\rho=(U_{r_1}-U_{r_\rho})/L$ is positive for tightening, where $r_1$ is the reference at $\rho=1$. Bounds and differences are in nats/token. The intervals are paired row-bootstrap 95\% intervals; they do not account for unknown numerical bias. PPL is the exponentiated estimated bound.}
 \begin{tabularx}{\linewidth}{@{}lrrrY@{}}
\toprule
$\rho$ & $\widehat U_\rho/L$ & Bound PPL & $\widehat\Delta_\rho$ & 95\% interval for $\Delta_\rho$ \\
\midrule
$1$ & 3.5477301 & 34.7344 & 0 & --- \\
$0.99$ & 3.6229472 & 37.4478 & $-0.0752171$ & $[-0.0759432,\ -0.0744884]$ \\
$0.9998$ & 3.5478568 & 34.7388 & $-0.0001267$ & $[-0.0002759,\ +0.0000222]$ \\
\bottomrule
\end{tabularx}

 \label{tab:likelihood}
\end{table}

Table~\ref{tab:likelihood} reports the held-out comparisons. At $\rho=0.99$, the estimated bound worsens by about $0.0752$ nats/token, raising bound PPL from $34.7344$ to $37.4478$. At the exploratory point $\rho=0.9998$, the paired improvement is $-1.267\times10^{-4}$ nats/token, with an interval spanning zero. Neither comparison provides evidence of positive tightening. The interval for this single exploratory point is not an upper bound on the best possible improvement over the entire reference family, and neither result supplies an additional upper bound on the total gap.

\paragraph{Numerical sensitivity.}
Matched comparisons on eight rows and two draws per row vary the solver steps or trace probes. For $\rho=0.9998$, changing from 128 to 256 steps shifts the estimated improvement by $-0.000696$ nats/token, with row-level standard error $0.000525$. These small controls do not certify accuracy at the $10^{-4}$ nats/token scale. The larger deterioration at $\rho=0.99$ remains visible across the checked settings. We therefore retain the negative result of the tested search, without interpreting a fitted near-baseline optimum as a measurable gain or interpreting the experiment as a measurement of true PPL.

\subsection{Controlled generation-metric audit}\label{sec:audit-setup}
The audit uses the first locally available OWT validation shard, containing 55,237 wrapped rows of 1024 tokens. Each selected source row contributes its first 128 GPT-2 tokens. Sampling without replacement produces a 512-row calibration set and three disjoint folds, each with 512 reference rows and a 2048-row candidate pool. Each evaluated output set contains 512 rows. Source rows do not overlap between these groups, but the cache lacks document identifiers, so document independence cannot be guaranteed.

GPT-2-large supplies both the judge likelihood and the text features. MAUVE uses FP32 last-token features, its default preprocessing, 51 clusters under the package's automatic setting, and three clustering seeds. The MAUVE pooling experiment uses matched preprocessing.

The constructions include fixed-size support reduction, judge-based selection, local token splicing, block reordering, whitespace changes, near-duplicates, and a numerical-consistency template. For support reduction, the first $k$ texts of the control set are repeated to keep the total output count at 512. For local splicing, 13 tokens are replaced using a separate donor row. Sparse versions apply that edit to only a fraction of the rows. These are controlled diagnostic constructions, not samples from newly trained CDLMs. Appendix~\ref{app:generation-protocol} supplies the detailed settings.

\subsection{Collapse is invisible to the conventional pair}\label{sec:stress-results}
\begin{table}[!htb]
 \centering\small
 \caption{\textbf{Controlled text constructions, 512 outputs per set.} Entries are means over three folds. For MAUVE, $\pm$ gives the standard deviation across folds. Within each fold, Gen PPL aggregates NLL over scored tokens. A low Gen PPL or high $\Hrow$ alone does not imply a better generator.}
 \begin{tabularx}{\linewidth}{@{}Yrrr@{}}
\toprule
Construction & Gen PPL & $H_{\mathrm{row}}$ & MAUVE \\
\midrule
Real control & 25.960 & 4.275 & $0.96026\pm0.00221$ \\
Selected text $\times512$ & 2.461 & 4.419 & $0.00407\pm0.00000$ \\
$256$ texts $\times2$ & 25.944 & 4.274 & $0.88665\pm0.02012$ \\
$64$ texts $\times8$ & 25.693 & 4.273 & $0.05743\pm0.01755$ \\
Judge-easiest quartile & 13.427 & 4.178 & $0.93391\pm0.01353$ \\
Middle splice, all rows & 44.248 & 4.312 & $0.95813\pm0.00421$ \\
Middle splice, 10\% rows & 27.411 & 4.278 & $0.96150\pm0.00938$ \\
Four-block reordering & 49.648 & 4.277 & $0.95417\pm0.00954$ \\
Doubled spaces & 14.311 & 3.185 & $0.07345\pm0.00595$ \\
\bottomrule
\end{tabularx}

 \label{tab:stress}
\end{table}

For the single-text counterexample, each fold selects a candidate whose unigram entropy is at least the control mean, then minimizes judge NLL within that eligible pool. Repeating this text 512 times reduces mean Gen PPL from $25.960$ to $2.461$ and increases $\Hrow$ from $4.275$ to $4.419$ (Table~\ref{tab:stress}). Nevertheless, its output distribution has one support point. The construction verifies the measurement blind spot without claiming that the selected text is the best text according to humans.
\FloatBarrier

MAUVE falls from $0.96026$ on the real control to $0.00407$ under single-text collapse. Reducing the support to 64 texts also lowers MAUVE while leaving Gen PPL and $\Hrow$ near their control values.

\begin{figure}[!htb]
 \centering
 \includegraphics[width=\linewidth]{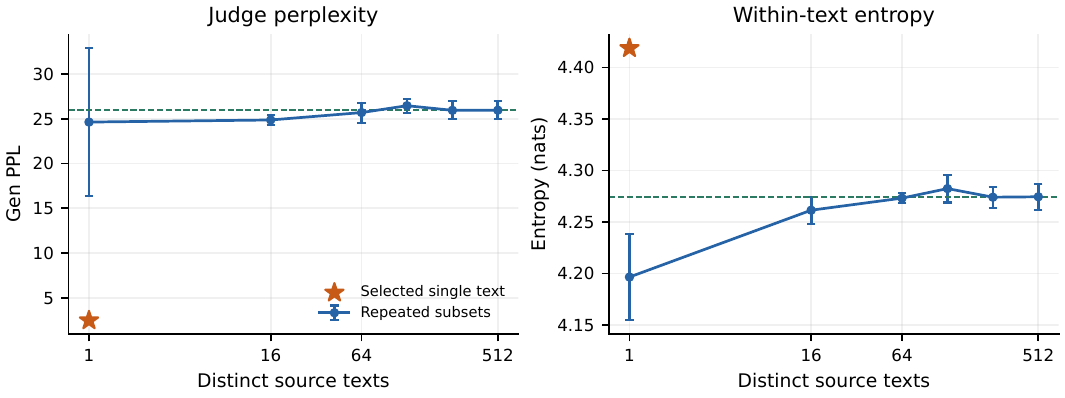}
 \caption{\textbf{Fixed output count, shrinking text support.} The control and repeated subsets all contain 512 rows. Points show fold means with fold standard deviations. The ordinary one-text subset uses the first control text; the star separately marks the low-NLL, high-entropy selection. The conventional pair does not consistently indicate loss of support. These deterministic constructions are descriptive stress tests, not additional independent draws.}
 \label{fig:support}
\end{figure}

Appending distinct numeric suffixes to repetitions of 64 texts makes the exact-duplicate rate zero.

\subsection{Detecting generation failures with KTD}\label{sec:failure-mode-metrics}
All five controlled failures in Table~\ref{tab:intro-failure-modes} yield lower Gen PPL than real OWT text.
Repeating one passage across all 1,024 outputs lowers Gen PPL from $14.91$ to $6.43$ and even raises sample entropy from $5.436$ to $5.735$ nats.
KTD assigns its largest discrepancies to exact duplication ($1.020$) and template collapse ($0.939$), compared with a near-zero score between two disjoint sets of real texts. MAUVE also detects both forms of collapse, falling to $0.004$ in each case.

\begin{table}[!htb]
 \centering\small
 \setlength{\tabcolsep}{4pt}
 \caption{\textbf{Generation metrics for the OpenWebText constructions illustrated in Section~\ref{sec:introduction}.}
 Each condition contains 1,024 texts of 1,024 GPT-2 tokens. MAUVE and KTD use a common, disjoint reference set of 1,024 OpenWebText texts.
 GPT-2-large supplies Gen PPL and the features for KTD and MAUVE. MAUVE uses FP32 last-token features and 102 clusters, reporting mean $\pm$ SD across three clustering seeds.
 $\Hrow$ (sample entropy) and $\JSDrows$ are in nats.}
 \label{tab:intro-failure-modes}
 \begin{tabularx}{\linewidth}{@{}Yrrrrr@{}}
\toprule
Condition & Gen PPL & $\Hrow$ & $\JSDrows$ & MAUVE & KTD \\
\midrule
Real OWT control & 14.91 & 5.436 & 2.078 & $0.962 \pm 0.005$ & 0.000 \\
\midrule
Exact duplication & 6.43 & 5.735 & 0.000 & $0.004 \pm 0.000$ & 1.020 \\
Within-text repetition & 4.62 & 5.263 & 2.444 & $0.052 \pm 0.006$ & 0.246 \\
Low-NLL stitching & 11.42 & 5.584 & 1.927 & $0.826 \pm 0.013$ & 0.043 \\
Template collapse & 5.54 & 5.242 & 0.209 & $0.004 \pm 0.000$ & 0.939 \\
Mixed degradations & 11.37 & 5.441 & 2.087 & $0.639 \pm 0.040$ & 0.029 \\
\bottomrule
\end{tabularx}

\end{table}
\FloatBarrier

The mixture of repetitive texts and stitched passages illustrates how averaging can mask two different failures.
Gen PPL falls to $11.37$, while sample entropy and $\JSDrows$ each stay within $0.01$ nats of the real-text control.
KTD still increases to $0.029$ for the mixture, with larger discrepancies for low-NLL stitching ($0.043$) and within-text repetition ($0.246$).
Across these constructions, KTD captures shifts in the distribution of text representations even when the conventional scores look favorable.

\subsection{Model and sampler comparisons}\label{sec:model-comparisons}
We compare recent discrete and continuous diffusion language models, all trained on OpenWebText. For methods without official checkpoints, we evaluate checkpoints from our own reproductions. We follow the original paper in reproducing RePlaid. In our reproduction, we use the Qwen3 tokenizer, whose vocabulary is substantially larger than that of the GPT-2-large tokenizer used by the other methods.

\noindent\begin{minipage}{\linewidth}
 \captionsetup{type=table}
 \centering\small
 \caption{\textbf{Evaluation results for discrete and continuous diffusion language models on OpenWebText.} \textsuperscript{*} The AR baseline comes from the DUO reproduction.}
 \label{tab:main-metrics}
 \begin{tabularx}{\linewidth}{@{}Yrrrrr>{\columncolor{black!7}}r}
\toprule
Methods & NFE & Gen PPL & $H_{rows}$ & $JSD_{rows}$ & MAUVE & \cellcolor{black!16}KTD \\
\midrule
Real validation set & - & 15.11 & 5.44 & 2.13 & 0.962 & 2e-6 \\
\midrule
\multicolumn{6}{@{}l}{\textbf{AR}} & \\
Transformer\textsuperscript{*} & 1024 & 35.36 & 5.58 & 2.01 & 0.932 & 0.017 \\
\midrule
\multicolumn{6}{@{}l}{\textbf{Discrete DLM}} & \\
MDLM & 1024 & 42.25 & 5.28 & 1.58 & 0.640 & 0.073 \\
SEDD & 1024 & 41.03 & 5.26 & 1.56 & 0.646 & 0.076 \\
DUO & 1024 & 77.18 & 5.54 & 1.93 & 0.855 & 0.056 \\
\midrule
\multicolumn{6}{@{}l}{\textbf{Continuous DLM}} & \\
ELF & 1024 & 7.30 & 4.74 & 2.29 & 0.006 & 0.088 \\
FLM & 1024 & 61.58 & 5.34 & 1.23 & 0.253 & 0.173 \\
LangFlow & 1024 & 36.16 & 5.25 & 2.15 & 0.900 & 0.027 \\
RePlaid$^{\dagger}$ (SDE) & 1024 & 17.09 & 5.28 & 2.10 & 0.685 & 0.028 \\
RePlaid$^{\dagger}$ (ODE) & 1024 & 22.47 & 5.33 & 2.07 & 0.620 & 0.026 \\
\bottomrule
\end{tabularx}

\end{minipage}

All three discrete baselines in Table~\ref{tab:main-metrics} have higher Gen PPL and lower $\JSDrows$ than the leading continuous DLMs in this comparison. Their $\JSDrows$ values range from 1.56 to 1.93, below the real-text value of 2.13, indicating less variation in token frequencies across samples. Overall, recent continuous DLMs compare favorably with the discrete baselines on both measures. ELF has the lowest Gen PPL (7.30), but its within-sequence entropy is lower than that of real text (4.74 versus 5.44). Among the diffusion models, RePlaid has the lowest KTD (0.026). For RePlaid, the SDE sampler yields lower Gen PPL than the ODE sampler (17.09 versus 22.47), while the ODE sampler has slightly lower KTD (0.026 versus 0.028). Despite their advantage in Gen PPL, the evaluated CDLMs still exhibit less lexical diversity within individual samples than real text does.

\subsection{Lexical diversity and vocabulary coverage}\label{sec:lexical-coverage}
The Gen PPL and sample entropy commonly reported for DLMs describe token predictability and diversity within individual texts. Vocabulary coverage asks which words appear across the entire collection.
For example, RePlaid (SDE) has lower Gen PPL than LangFlow ($17.09$ versus $36.16$) at similar sample entropy ($5.28$ versus $5.25$) in Table~\ref{tab:main-metrics}.
Yet at a common budget of 705,115 word occurrences, RePlaid (SDE) yields fewer distinct word forms in expectation ($37{,}384$ versus $41{,}763$, Table~\ref{tab:lexical-diversity}).

\begin{table}[!htb]
 \centering\small
 \setlength{\tabcolsep}{3pt}
 \caption{\textbf{Vocabulary statistics for OWT and model-generated text.} Each model row and OWT (1,024) use 1,024 contexts sampled uniformly without replacement with seed 20261120. OWT (full) contains all 110,473 validation contexts. Exported texts retain their available lengths. Words contain at least one letter, exclude URLs and email addresses, and use NFKC normalization and case folding. Special-token markers delimit paragraphs. Types are distinct word forms and lemmas are spaCy base forms. $V_n$ is the expected type count in a uniform draw of $n=705{,}115$ word occurrences without replacement. $C_{1k}$ and $C_{\rm full}$ give the percentages of the sampled and full OWT vocabularies covered by each row.}
 \label{tab:lexical-diversity}
\begin{tabularx}{\linewidth}{@{}Yrrrrrrr@{}}
\toprule
Source & Contexts & \shortstack{Word\\occurrences} & Types & Lemmas & $V_n$ & $C_{1k}$ & $C_{\rm full}$ \\
\midrule
OWT (full) & 110,473 & 82,075,702 & 713,028 & 695,324 & 50,732 & 100.00 & 100.00 \\
OWT (1,024) & 1,024 & 758,171 & 44,019 & 37,450 & 42,590 & 100.00 & 6.17 \\
\midrule
AR (DUO) & 1,024 & 766,321 & 49,477 & 42,845 & 47,381 & 54.25 & 5.69 \\
MDLM & 1,024 & 796,466 & 29,405 & 24,707 & 27,687 & 40.36 & 3.40 \\
SEDD & 1,024 & 802,562 & 29,153 & 24,430 & 27,345 & 39.90 & 3.32 \\
DUO & 1,024 & 763,542 & 48,998 & 42,576 & 46,842 & 51.33 & 4.95 \\
ELF & 1,024 & 735,131 & 17,265 & 15,157 & 17,034 & 19.66 & 1.55 \\
FLM & 1,024 & 782,665 & 28,258 & 24,760 & 26,560 & 31.53 & 2.71 \\
LangFlow & 1,024 & 731,324 & 42,580 & 37,386 & 41,763 & 44.24 & 3.80 \\
RePlaid (SDE) & 1,024 & 709,886 & 37,492 & 33,519 & 37,384 & 40.78 & 3.83 \\
RePlaid (ODE) & 1,024 & 705,115 & 42,009 & 37,687 & 42,009 & 43.12 & 4.12 \\
\bottomrule
\end{tabularx}

\end{table}

At 1,024 contexts per model, every evaluated DLM covers less of the sampled OWT vocabulary than the AR Transformer. Coverage ranges from $19.66\%$ for ELF to $51.33\%$ for DUO, compared with $54.25\%$ for AR.
DUO also comes closest to AR on the full OWT validation vocabulary ($4.95\%$ versus $5.69\%$).

\begin{table}[!htb]
 \centering\footnotesize
 \setlength{\tabcolsep}{2pt}
 \renewcommand{\arraystretch}{1.05}
 \caption{\textbf{Part-of-speech distributions for OWT and model-generated text.} Panel A reports percentages of word occurrences and Panel B percentages of distinct (word form, POS) pairs, using the contexts and word rules of Table~\ref{tab:lexical-diversity}. Tags and lemmas use spaCy 3.7.5 with \texttt{en\_core\_web\_sm} 3.7.1. A form used with multiple POS tags contributes one type to each corresponding category. Articles comprise \textit{a}, \textit{an}, and \textit{the} tagged as determiners, and the determiners row contains the remaining determiners. Other combines the remaining X, SYM, PUNCT, and SPACE tags. All entries are percentages.}
 \label{tab:pos-distribution}
\begin{tabular*}{\linewidth}{@{\extracolsep{\fill}}l*{11}{r}@{}}
\toprule
\multicolumn{12}{@{}l}{\textbf{A. Word-occurrence shares}} \\
POS & \shortstack{OWT\\full} & \shortstack{OWT\\1,024} & AR & MDLM & SEDD & DUO & ELF & FLM & LangFlow & \shortstack{RePlaid\\SDE} & \shortstack{RePlaid\\ODE} \\
\midrule
Common nouns & 20.75 & 20.82 & 20.72 & 17.76 & 17.69 & 20.90 & 20.98 & 17.59 & 20.10 & 19.42 & 19.50 \\
Proper nouns & 10.94 & 10.92 & 11.24 & 6.42 & 6.00 & 10.10 & 12.86 & 5.52 & 12.17 & 14.53 & 14.85 \\
Verbs & 12.43 & 12.41 & 12.25 & 13.88 & 13.88 & 12.62 & 12.68 & 13.71 & 12.42 & 10.90 & 10.87 \\
Auxiliaries & 5.73 & 5.71 & 5.44 & 7.17 & 7.22 & 5.85 & 5.31 & 6.98 & 5.82 & 5.63 & 5.49 \\
Adjectives & 7.44 & 7.48 & 7.60 & 6.42 & 6.53 & 7.40 & 7.01 & 6.29 & 7.10 & 7.06 & 7.01 \\
Adverbs & 4.39 & 4.39 & 4.41 & 5.34 & 5.43 & 4.51 & 2.73 & 5.60 & 4.13 & 3.76 & 3.75 \\
Articles & 8.31 & 8.30 & 8.23 & 7.68 & 7.71 & 8.22 & 9.87 & 8.11 & 8.20 & 8.85 & 8.82 \\
Determiners & 1.30 & 1.30 & 1.28 & 1.54 & 1.55 & 1.35 & 0.93 & 1.48 & 1.25 & 1.12 & 1.07 \\
Pronouns & 7.56 & 7.50 & 7.45 & 12.10 & 12.13 & 7.91 & 6.49 & 12.16 & 7.73 & 7.04 & 7.04 \\
Prepositions & 11.70 & 11.71 & 11.98 & 10.75 & 10.86 & 11.67 & 12.84 & 11.62 & 11.71 & 12.55 & 12.48 \\
Coord.\ conj. & 3.32 & 3.31 & 3.32 & 3.51 & 3.49 & 3.31 & 2.41 & 3.44 & 3.21 & 3.47 & 3.49 \\
Subord.\ conj. & 2.14 & 2.14 & 2.07 & 2.73 & 2.77 & 2.13 & 1.54 & 2.66 & 2.04 & 1.70 & 1.67 \\
Particles & 3.09 & 3.11 & 3.12 & 3.71 & 3.77 & 3.07 & 3.35 & 3.77 & 3.12 & 2.60 & 2.63 \\
Numerals & 0.71 & 0.68 & 0.70 & 0.71 & 0.71 & 0.72 & 0.82 & 0.74 & 0.78 & 1.11 & 1.05 \\
Interjections & 0.11 & 0.11 & 0.10 & 0.21 & 0.20 & 0.11 & 0.14 & 0.27 & 0.10 & 0.08 & 0.09 \\
Other & 0.11 & 0.11 & 0.09 & 0.07 & 0.06 & 0.11 & 0.04 & 0.07 & 0.11 & 0.18 & 0.19 \\
\midrule
\multicolumn{12}{@{}l}{\textbf{B. Distinct word--POS shares}} \\
POS & \shortstack{OWT\\full} & \shortstack{OWT\\1,024} & AR & MDLM & SEDD & DUO & ELF & FLM & LangFlow & \shortstack{RePlaid\\SDE} & \shortstack{RePlaid\\ODE} \\
\midrule
Common nouns & 24.60 & 31.12 & 30.15 & 32.07 & 32.23 & 30.72 & 28.14 & 32.48 & 29.10 & 27.77 & 27.95 \\
Proper nouns & 52.46 & 37.60 & 40.46 & 34.04 & 33.88 & 39.65 & 42.06 & 33.40 & 40.52 & 43.38 & 43.92 \\
Verbs & 7.24 & 15.99 & 14.99 & 17.76 & 17.97 & 14.95 & 16.32 & 17.70 & 15.67 & 13.51 & 12.89 \\
Auxiliaries & 0.21 & 0.20 & 0.17 & 0.32 & 0.31 & 0.21 & 0.33 & 0.49 & 0.19 & 0.15 & 0.15 \\
Adjectives & 7.26 & 10.53 & 9.90 & 10.21 & 10.38 & 9.68 & 8.21 & 10.10 & 9.59 & 9.45 & 9.33 \\
Adverbs & 1.51 & 2.42 & 2.25 & 2.77 & 2.61 & 2.17 & 2.33 & 2.99 & 2.33 & 2.06 & 1.97 \\
Articles & 0.00034 & 0.00545 & 0.00491 & 0.00839 & 0.00850 & 0.00504 & 0.01436 & 0.00899 & 0.00590 & 0.00669 & 0.00600 \\
Determiners & 0.09 & 0.06 & 0.06 & 0.09 & 0.10 & 0.07 & 0.12 & 0.12 & 0.07 & 0.07 & 0.07 \\
Pronouns & 0.30 & 0.23 & 0.20 & 0.34 & 0.31 & 0.22 & 0.38 & 0.41 & 0.21 & 0.20 & 0.23 \\
Prepositions & 0.40 & 0.27 & 0.25 & 0.31 & 0.30 & 0.29 & 0.41 & 0.38 & 0.28 & 0.31 & 0.30 \\
Coord.\ conj. & 0.06 & 0.04 & 0.03 & 0.04 & 0.05 & 0.04 & 0.09 & 0.06 & 0.04 & 0.05 & 0.04 \\
Subord.\ conj. & 0.04 & 0.10 & 0.09 & 0.15 & 0.13 & 0.10 & 0.20 & 0.16 & 0.09 & 0.10 & 0.09 \\
Particles & 0.02 & 0.02 & 0.02 & 0.03 & 0.03 & 0.02 & 0.05 & 0.03 & 0.03 & 0.02 & 0.03 \\
Numerals & 2.44 & 0.59 & 0.60 & 0.72 & 0.64 & 0.83 & 0.59 & 0.74 & 0.80 & 1.30 & 1.34 \\
Interjections & 0.25 & 0.12 & 0.12 & 0.21 & 0.17 & 0.16 & 0.18 & 0.24 & 0.16 & 0.15 & 0.12 \\
Other & 3.11 & 0.72 & 0.71 & 0.91 & 0.88 & 0.89 & 0.59 & 0.70 & 0.92 & 1.47 & 1.58 \\
\bottomrule
\end{tabular*}

\end{table}
\FloatBarrier

Table~\ref{tab:pos-distribution} shows that this coverage gap takes different forms across models.
Proper nouns account for $10.92\%$ of word occurrences in the 1,024-context OWT sample and $11.24\%$ in AR outputs, compared with $5.52$--$6.42\%$ in FLM, SEDD, and MDLM.
Proper nouns also make up a smaller share of distinct word--POS pairs in those three models ($33.40$--$34.04\%$) than in OWT ($37.60\%$).
For both RePlaid samplers that share exceeds $43\%$, illustrating how a model can favor proper nouns while covering less of the OWT vocabulary overall.

\section{Related Work}\label{sec:related}
\paragraph{Continuous diffusion language modeling.}
Plaid develops likelihood-based diffusion language modeling, LangFlow studies continuous language generation through flow-based modeling, and RePlaid revisits continuous diffusion at scale \citep{gulrajani2023plaid,chen2026langflow,yang2026replaid}. ELF supplies another recent continuous-space formulation and emphasizes generation evaluation \citep{hu2026elf}. These systems motivate a benchmark that can accommodate different likelihood interfaces and sampling procedures. Our analysis does not assume that every CDLM reports the same metrics: RePlaid, for example, also reports MAUVE. The collapse argument applies specifically to reliance on the Gen PPL and within-sequence entropy pair.

\paragraph{Likelihood bounds and density estimation.}
Variational diffusion supplies tractable objectives for latent generative models \citep{kingma2021vdm}; learned inference or dequantization references and importance-weighted bounds provide ways to tighten likelihood estimates \citep{ho2019flowpp,burda2016iwae}. Continuous normalizing flows connect likelihood to ODE divergence integration \citep{chen2018neuralode,grathwohl2019ffjord}. We build on these foundations. Our analysis applies the established bounds to CDLM comparison rules and gap identifiability, and examines how stochastic log-density evaluation interacts with nonlinear importance weighting. The ELBO and importance-weighted bounds themselves are standard results.

\paragraph{Distributional generation metrics.}
MMD provides a kernel two-sample framework \citep{gretton2012kernel}. MAUVE evaluates text distributions through a divergence-frontier construction \citep{pillutla2021mauve}, while precision, recall, density, and coverage offer complementary generative-model diagnostics \citep{naeem2020reliable}. We adopt these established tools within a CDLM-specific evaluation protocol.

\section{Discussion and Limitations}\label{sec:discussion}
The benchmark separates questions that are often compressed into a single comparison: the probability assigned by an ideal model, the tightness of its likelihood bound, the behavior of a practical sampler, and the discrepancy visible to a chosen measurement. This separation makes an improvement easier to interpret. It also requires retaining trade-offs when different measurements disagree.

The empirical scope is limited. The likelihood study covers one archived LangFlow checkpoint and a one-parameter reference family; it does not measure RePlaid's gap or either model's exact PPL. The generation study is a short-text OWT pilot with three folds from one available shard and no document identifiers. Its edited texts and selected subsets isolate specific evaluation failure modes. Table~\ref{tab:stress} uses MAUVE with its default last-token features.

Interpreting a benchmark comparison requires faithful model implementations, a specified output distribution, and an evaluation protocol with an appropriate independent sampling unit. Runtime comparisons require accounting for representation extraction as well as discrepancy computation. A single score cannot settle an unspecified trade-off between corpus fidelity, human preference, rare errors, and computational cost. Explicit task requirements and complementary measurements make those decisions reviewable.

\section{Conclusion}\label{sec:conclusion}
Reliable CDLM development requires measurements with clear targets and reproducible experimental objects. We characterize the limits of comparing likelihood bounds, show why a reference-scale optimum leaves the total gap unidentified, and distinguish ideal multi-sample gap reduction from effects of stochastic density evaluation. We also demonstrate that Gen PPL with within-sequence entropy can favor an entirely collapsed output distribution. The MAUVE experiments show how sensitivity to perturbations depends on the text representation. Open CDLM connects these findings to a unified evaluation protocol and a library design for unofficial LangFlow and RePlaid reproductions.

\par\bigskip
\begingroup
\small
\setlength{\bibsep}{2pt}
\bibliographystyle{plainnat}
\bibliography{references,paper_references}

@article{gretton2012kernel,
  title={A Kernel Two-Sample Test},
  author={Gretton, Arthur and Borgwardt, Karsten M. and Rasch, Malte J. and Sch{\"o}lkopf, Bernhard and Smola, Alexander},
  journal={Journal of Machine Learning Research},
  volume={13},
  number={25},
  pages={723--773},
  year={2012},
  url={https://www.jmlr.org/papers/v13/gretton12a.html}
}

@inproceedings{pillutla2021mauve,
  title={{MAUVE}: Measuring the Gap Between Neural Text and Human Text using Divergence Frontiers},
  author={Pillutla, Krishna and Swayamdipta, Swabha and Zellers, Rowan and Thickstun, John and Welleck, Sean and Choi, Yejin and Harchaoui, Zaid},
  booktitle={Advances in Neural Information Processing Systems},
  year={2021},
  url={https://arxiv.org/abs/2102.01454}
}

@article{gulrajani2023plaid,
  title={Likelihood-Based Diffusion Language Models},
  author={Gulrajani, Ishaan and Hashimoto, Tatsunori B.},
  journal={arXiv preprint arXiv:2305.18619},
  year={2023},
  url={https://arxiv.org/abs/2305.18619}
}

@article{hu2026elf,
  title={{ELF}: Embedded Language Flows},
  author={Hu, Keya and Qiu, Linlu and Lu, Yiyang and Zhao, Hanhong and Li, Tianhong and Kim, Yoon and Andreas, Jacob and He, Kaiming},
  journal={arXiv preprint arXiv:2605.10938},
  year={2026},
  note={Version 2; Section 4 describes the evaluation protocol},
  url={https://arxiv.org/html/2605.10938v2}
}

@misc{langflowrepo2026,
  title={{LangFlow}: Official Repository},
  author={{LangFlow authors}},
  year={2026},
  howpublished={GitHub repository},
  note={README inspected on 6 November 2026},
  url={https://github.com/nealchen2003/LangFlow}
}

@inproceedings{naeem2020reliable,
  title={Reliable Fidelity and Diversity Metrics for Generative Models},
  author={Naeem, Muhammad Ferjad and Oh, Seong Joon and Uh, Youngjung and Choi, Yunjey and Yoo, Jaejun},
  booktitle={Proceedings of the 37th International Conference on Machine Learning},
  year={2020},
  url={https://arxiv.org/abs/2002.09797}
}

@article{chen2026langflow,
  title={LangFlow: Continuous Diffusion Rivals Discrete in Language Modeling},
  author={Chen, Yuxin and Liang, Chumeng and Sui, Hangke and Guo, Ruihan and Cheng, Chaoran and You, Jiaxuan and Liu, Ge},
  journal={arXiv preprint arXiv:2604.11748},
  year={2026},
  note={Version 3. Accessed 2026-11-02},
  url={https://arxiv.org/html/2604.11748v3}
}

@article{yang2026replaid,
  title={Continuous Diffusion Scales Competitively with Discrete Diffusion for Language},
  author={Yang, Zhihan and Guo, Wei and Zhang, Shuibai and Sahoo, Subham Sekhar and Chen, Yongxin and Vahdat, Arash and Mardani, Morteza and Thickstun, John},
  journal={arXiv preprint arXiv:2605.18530},
  year={2026},
  note={Version 3; see Appendix F for ODE/IWAE evaluation. Accessed 2026-11-02},
  url={https://arxiv.org/html/2605.18530v3}
}

@misc{gokaslan2019openwebtext,
  title={OpenWebText Corpus},
  author={Gokaslan, Aaron and Cohen, Vanya and Pavlick, Ellie and Tellex, Stefanie},
  year={2019},
  howpublished={\url{https://skylion007.github.io/OpenWebTextCorpus}}
}

@inproceedings{kingma2021vdm,
  title={Variational Diffusion Models},
  author={Kingma, Diederik P. and Salimans, Tim and Poole, Ben and Ho, Jonathan},
  booktitle={Advances in Neural Information Processing Systems},
  year={2021},
  url={https://arxiv.org/abs/2107.00630}
}

@inproceedings{ho2019flowpp,
  title={Flow++: Improving Flow-Based Generative Models with Variational Dequantization and Architecture Design},
  author={Ho, Jonathan and Chen, Xi and Srinivas, Aravind and Duan, Yan and Abbeel, Pieter},
  booktitle={International Conference on Machine Learning},
  year={2019},
  url={https://arxiv.org/abs/1902.00275}
}

@inproceedings{chen2018neuralode,
  title={Neural Ordinary Differential Equations},
  author={Chen, Ricky T. Q. and Rubanova, Yulia and Bettencourt, Jesse and Duvenaud, David},
  booktitle={Advances in Neural Information Processing Systems},
  year={2018},
  url={https://arxiv.org/abs/1806.07366}
}

@article{luo2026designing,
  title={Designing Transformers for Continuous Diffusion Language Models},
  author={Luo, Weijian and Chen, Yujian and Jiang, Hao and Cheng, Guang and Hu, Tianyang},
  journal={arXiv preprint arXiv:2610.22330},
  year={2026}
}

@inproceedings{song2021sde,
  title={Score-Based Generative Modeling through Stochastic Differential Equations},
  author={Song, Yang and Sohl-Dickstein, Jascha and Kingma, Diederik P. and Kumar, Abhishek and Ermon, Stefano and Poole, Ben},
  booktitle={International Conference on Learning Representations},
  year={2021},
  url={https://arxiv.org/abs/2011.13456}
}

@inproceedings{burda2016iwae,
  title={Importance Weighted Autoencoders},
  author={Burda, Yuri and Grosse, Roger and Salakhutdinov, Ruslan},
  booktitle={International Conference on Learning Representations},
  year={2016},
  url={https://arxiv.org/abs/1509.00519}
}

@article{grosse2015bdmc,
  title={Sandwiching the Marginal Likelihood Using Bidirectional Monte Carlo},
  author={Grosse, Roger B. and Ghahramani, Zoubin and Adams, Ryan P.},
  journal={arXiv preprint arXiv:1511.02543},
  year={2015},
  url={https://arxiv.org/abs/1511.02543}
}

@inproceedings{grathwohl2019ffjord,
  title={{FFJORD}: Free-form Continuous Dynamics for Scalable Reversible Generative Models},
  author={Grathwohl, Will and Chen, Ricky T. Q. and Bettencourt, Jesse and Sutskever, Ilya and Duvenaud, David},
  booktitle={International Conference on Learning Representations},
  year={2019},
  url={https://arxiv.org/abs/1810.01367}
}
\endgroup
\clearpage
\appendix
\section{Likelihood Identities and Counterexamples}\label{app:likelihood-proofs}
\subsection{Continuous density and discrete probability}\label{app:density}
Let $q(x,z)=P(x)r(z\mid x)$ and $p_\theta(x,z)=p_\theta(z)c_\theta(x\mid z)$. Write $q_Z$ for the marginal of $q$. Applying the chain rule to the same joint KL in two orders gives
\begin{align}
 \KL(q\Vert p_\theta)
 &=\KL(P\Vert P_\theta)+
   \E_{P}\KL(r(z\mid X)\Vert p_\theta(z\mid X)) \notag\\
 &=\KL(q_Z\Vert p_\theta(z))+
   \E_{q_Z}\KL(q(x\mid Z)\Vert c_\theta(x\mid Z)).
 \label{eq:joint-kl}
\end{align}
Adding the discrete entropy $H(P)$ produces
\begin{align}
 \E_P U_r(X)
 &=H(P)+\KL(P\Vert P_\theta)+\E_PG_r(X) \notag\\
 &=H(P)+\KL(q_Z\Vert p_\theta(z))
   +\E_{q_Z}\KL(q(x\mid Z)\Vert c_\theta(x\mid Z)).
 \label{eq:joint-views}
\end{align}
This separates discrete model fit and variational slack while showing how continuous density fit and decoder error contribute to the same bound.

For completeness, let $C_Z=\E_{q_Z}[-\log p_\theta(Z)]$ be continuous cross-entropy and $h$ denote differential entropy. Then
\begin{align}
 h(Z)&=H(X)+h(Z\mid X)-H(X\mid Z),\notag\\
 C_Z-h(Z\mid X)
 &=H(X)-H(X\mid Z)+\KL(q_Z\Vert p_\theta(z)).
 \label{eq:entropy-correction}
\end{align}
Subtracting a conditional Gaussian entropy alone does not generally produce discrete text cross-entropy: the conditional discrete entropy remains. The decoder cross-entropy contributes the corresponding $H(X\mid Z)$ and its own conditional KL. Explicitly,
\begin{equation}
 \E_P U_r(X)=C_Z-h(Z\mid X)+\E_{(X,Z)\sim q}[-\log c_\theta(X\mid Z)].
 \label{eq:density-decoder-bound}
\end{equation}
The valid latent-model bound includes all these terms rather than equating a density at an embedding with a text probability.

\paragraph{Small-noise peaks and Gaussian entropy.}
Consider a finite collection of labels with probabilities $P(k)$ and Gaussian corruption $r_\tau(z\mid k)=\mathcal N(\alpha_\tau e_k,\sigma_\tau^2I_D)$. Its marginal $q_{Z,\tau}(z)=\sum_kP(k)r_\tau(z\mid k)$ obeys
\begin{align}
 q_{Z,\tau}(\alpha_\tau e_k)&\geq P(k)(2\pi\sigma_\tau^2)^{-D/2},\notag\\
 -\sum_kP(k)\log q_{Z,\tau}(\alpha_\tau e_k)
 &\leq H(P)+\frac D2\log(2\pi\sigma_\tau^2).
 \label{eq:gaussian-center-density}
\end{align}
The density at a center can diverge as $\sigma_\tau\to0$ although every discrete probability remains at most one. In contrast, the expected negative log-density of a single Gaussian component is
\begin{equation}
 h(Z\mid X)=\frac D2\log(2\pi e\sigma_\tau^2),
 \label{eq:gaussian-reference-entropy}
\end{equation}
which exceeds its center NLL by $D/2$. Center evaluation and noise-averaged evaluation therefore require different constants. Overlap between components contributes the remaining $H(X\mid Z)$ in \eqref{eq:entropy-correction}. A non-Gaussian model posterior, small-noise density divergence, and variational slack are distinct phenomena; even two Gaussians can have a large posterior KL.

\subsection{Continuous flows and decoder-defined probabilities}\label{app:flow-decoder}
For a smooth invertible flow $\dot z_t=v_\theta(z_t,t)$, the instantaneous change-of-variables identity is
\begin{equation}
 \frac{\mathrm d}{\mathrm dt}\log p_t(z_t)=-\nabla\!\cdot v_\theta(z_t,t),\qquad
 \log p_0(z_0)=\log p_T(z_T)+\int_0^T\nabla\!\cdot v_\theta(z_t,t)\,\mathrm dt.
 \label{eq:flow-density-identity}
\end{equation}
This computes a continuous density \citep{chen2018neuralode}. Under $y=az$ with $a\ne0$, the transformed density is $p_Y(y)=p_Z(y/a)/|a|^D$, so its differential NLL changes by $D\log|a|$. If the reference and decoder are transformed consistently, the same Jacobian enters the reference density and cancels in $p_\theta(z)c_\theta(x\mid z)/r(z\mid x)$. The discrete marginal and its bound are invariant to this coordinate change.

A deterministic decoder $d(z)$ defines cells $C_x=\{z:d(z)=x\}$ and a different discrete model,
\begin{align}
 P_\theta^{\mathrm{hard}}(x)&=\int_{C_x}p_\theta(z)\,\mathrm dz,\notag\\
 \E_{r_x}[\log r_x(Z)-\log p_\theta(Z)]
 &=-\log P_\theta^{\mathrm{hard}}(x)
 +\KL\!\left(r_x\Vert p_\theta(\cdot\mid C_x)\right).
 \label{eq:hard-decoder-bound}
\end{align}
Here $r_x$ must be normalized and supported inside $C_x$, with finite KL. Truncating a full-space reference $q$ requires $r_x(z)=q(z)\mathbf1_{C_x}(z)/Z_x$ and $Z_x=\int_{C_x}q(z)\,\mathrm dz$. Alternatively, a full-space importance estimator must retain the indicator $\mathbf1_{C_x}$ in its weight. A full-space Gaussian cannot be substituted into \eqref{eq:hard-decoder-bound} without these changes.

For a stochastic categorical decoder, $c_\theta(x\mid z)=\prod_{\ell=1}^{L_x}c_\theta(x_\ell\mid z)$ is a normalized conditional probability; each factor may depend on the entire latent sequence. Integrating over a shared $z$ generally does not preserve factorization across tokens. Replacing categorical sampling by argmax changes the decoder and can change the text marginal, even with identical continuous dynamics.

\subsection{Endpoint and path-space variational bounds}\label{app:path-bound}
For a finite diffusion path with $K$ transitions, write
\begin{equation}
 p_\theta^{\mathrm{path}}(x,z_{0:K})
 =c_\theta(x\mid z_0)p_K(z_K)
   \prod_{k=1}^Kp_\theta(z_{k-1}\mid z_k).
 \label{eq:path-joint-model}
\end{equation}
Let $q(z_{0:K}\mid x)$ be a forward Markov reference, and let $q_k$ denote its time-$k$ marginal. Factoring this reference in reverse time and taking expectations gives
\begin{align}
 U_{\mathrm{path}}(x)
 &=\E_{q_0}[-\log c_\theta(x\mid Z_0)]
 +\KL(q_K(\cdot\mid x)\Vert p_K)\notag\\
 &\quad+\sum_{k=1}^K\E_{Z_k\sim q_k(\cdot\mid x)}
 \KL\!\left(q(z_{k-1}\mid Z_k,x)\Vert p_\theta(z_{k-1}\mid Z_k)\right)\notag\\
 &=-\log P_\theta^{\mathrm{path}}(x)
 +\KL\!\left(q(z_{0:K}\mid x)\Vert p_\theta^{\mathrm{path}}(z_{0:K}\mid x)\right).
 \label{eq:path-nelbo}
\end{align}
The chain rule further decomposes this gap into
\begin{align}
 G_{\mathrm{path}}(x)
 &=\KL(q(z_0\mid x)\Vert p_\theta^{\mathrm{path}}(z_0\mid x))\notag\\
 &\quad+\E_{q(z_0\mid x)}
 \KL(q(z_{1:K}\mid z_0,x)\Vert p_\theta^{\mathrm{path}}(z_{1:K}\mid z_0,x)).
 \label{eq:path-gap-chain}
\end{align}
Marginalizing the rest of the path therefore cannot increase the ideal bound within this same joint model. Evaluating the resulting endpoint density may itself be intractable. This identity does not identify that endpoint marginal with a separately defined approximate ODE model.

For the standard variance-preserving Gaussian parameterization, let $q_\gamma(z\mid x)=\mathcal N(\alpha_\gamma e(x),\sigma_\gamma^2I)$, $\alpha_\gamma^2+\sigma_\gamma^2=1$, and $\gamma=\log(\sigma_\gamma^2/\alpha_\gamma^2)$. With the usual denoising parameterization of the reverse Gaussian mean and its matching conditional covariance, each transition KL reduces to a squared denoising error weighted by the decrease in signal-to-noise ratio. In the continuous-time limit, $\operatorname{SNR}=e^{-\gamma}$ yields the VDM objective \citep{kingma2021vdm}
\begin{align}
 U_{\mathrm{VDM}}(x)
 &=\E_{q_{\gamma_0}}[-\log c_\theta(x\mid Z)]
 +\KL(q_{\gamma_T}(\cdot\mid x)\Vert\mathcal N(0,I))\notag\\
 &\quad+\frac12\int_{\gamma_0}^{\gamma_T}e^{-\gamma}
 \E_{Z\sim q_\gamma(\cdot\mid x)}
 \|e(x)-\widehat e_\theta(Z,\gamma)\|^2\,\mathrm d\gamma.
 \label{eq:vdm-continuous-objective}
\end{align}
This expression requires the stated parameterization and boundary terms; arbitrary finite-step reverse transitions instead require the KL terms in \eqref{eq:path-nelbo}. A known Gaussian forward reference is not a claim that the learned marginal or posterior is Gaussian. The path-space version of \eqref{eq:importance-identity} uses the entire path as its latent variable. An unbiased Monte Carlo estimate of the time-integrated loss is not, by itself, a pointwise log joint/reference ratio that can be exponentiated to form an importance weight. The distinction between ODE and reverse-SDE model marginals remains subject to the score and boundary conditions discussed in Section~\ref{sec:likelihood}.

\subsection{Proofs for multi-sample bounds and gaps}\label{app:multisample-proof}
Fix $x$ and abbreviate $p_\theta(z\mid x)$ by $\pi_x(z)$ and $r(z\mid x)$ by $r_x(z)$. Assume $P_\theta(x)>0$ and mutual absolute continuity of these two densities. The condition $\pi_x\ll r_x$ ensures that no posterior mass is missed by importance sampling, while $r_x\ll\pi_x$, together with integrability, permits a finite single-sample NELBO. Direct integration gives
\begin{equation}
 \E_{r_x}w_r(Z,x)
 =\int r_x(z)\frac{p_\theta(z)c_\theta(x\mid z)}{r_x(z)}\,\mathrm dz
 =P_\theta(x),\qquad
 \frac{w_r(z,x)}{P_\theta(x)}=\frac{\pi_x(z)}{r_x(z)}.
 \label{eq:importance-proof}
\end{equation}
In particular,
\begin{equation}
 U_r(x)=-\log P_\theta(x)+\E_{r_x}\log\frac{r_x(Z)}{\pi_x(Z)},
 \label{eq:single-gap-proof}
\end{equation}
which proves \eqref{eq:gap}. If a proposal misses posterior support, its mean importance weight instead integrates only the covered mass; the convergence claim for the full marginal then does not apply.

\begin{proof}[Proof of Proposition~\ref{prop:multisample}]
Let $W_n=w_r(Z_n,x)$ and $\overline W_N=N^{-1}\sum_nW_n$. Jensen's inequality gives $-\E\log\overline W_N\geq-\log\E\overline W_N=-\log P_\theta(x)$. Given $N+1$ independent weights, define each leave-one-out mean by $A_j=N^{-1}\sum_{i\ne j}W_i$. Since $(N+1)^{-1}\sum_j A_j=\overline W_{N+1}$, concavity of the logarithm gives
\begin{equation}
 \log\overline W_{N+1}\geq\frac1{N+1}\sum_{j=1}^{N+1}\log A_j.
 \label{eq:iwae-monotonic-proof}
\end{equation}
Every $A_j$ has the distribution of $\overline W_N$. Taking expectations and changing sign proves $U_{r,N+1}\leq U_{r,N}$. The strong law gives $\overline W_N\to P_\theta(x)$ almost surely. Uniform integrability of $\log\overline W_N$ permits convergence of its expectation; the sufficient two-sided bound in the proposition implies this condition by bounded convergence. Almost-sure convergence alone does not justify exchanging the logarithmic expectation and the limit.
\end{proof}

The finite-$N$ gap has an exact KL representation on the product latent space. Define the normalized densities
\begin{align}
 R_{r,N}(z_{1:N}\mid x)&=\prod_{n=1}^N r_x(z_n),\notag\\
 T_{\theta,r,N}(z_{1:N}\mid x)&=\frac1N\sum_{n=1}^N
 \pi_x(z_n)\prod_{j\ne n}r_x(z_j).
 \label{eq:extended-densities}
\end{align}
The second density chooses one position uniformly for a posterior draw and fills the remaining positions from the reference. Their ratio and KL are
\begin{align}
 \frac{T_{\theta,r,N}}{R_{r,N}}
 &=\frac1N\sum_{n=1}^N\frac{\pi_x(z_n)}{r_x(z_n)}
 =\frac{\overline W_N}{P_\theta(x)},\notag\\
 G_{r,N}(x)&=\KL(R_{r,N}\Vert T_{\theta,r,N}).
 \label{eq:extended-gap-kl}
\end{align}
For $N=1$ this is the original posterior KL. For larger $N$ it compares different joint distributions, so retaining the single-sample KL as the remaining gap would be incorrect. If $r_x=\pi_x$, the weights are constant and all bounds are exact. Conversely, equality in the Jensen bound at finite $N$ requires an almost-surely constant sample-mean weight; independence then implies constant individual weights and $r_x=\pi_x$ almost everywhere.

\subsection{Sequence aggregation and empirical monotonicity}\label{app:weight-aggregation}
For log-weights $\ell_n(x)=\log w_r(Z_n,x)$, a single replicate of the multi-sample estimator is
\begin{equation}
 \widehat U_{r,N}(x)
 =-\operatorname{LSE}(\ell_1(x),\ldots,\ell_N(x))+\log N,
 \qquad \operatorname{LSE}(a_{1:N})=\log\sum_{n=1}^N e^{a_n}.
 \label{eq:sequence-logsumexp}
\end{equation}
Independent replicates estimate $U_{r,N}$. Corpus reporting first sums these sequence estimates and then divides by the total scored tokens as in \eqref{eq:corpus-ppl}. Moving token normalization inside the exponential replaces $w_n$ by $w_n^{1/L_x}$ and changes the objective. Even when the decoder factorizes, in general
\begin{equation}
 \frac1N\sum_{n=1}^N\prod_{\ell=1}^{L_x}c_\theta(x_\ell\mid Z_n)
 \ne\prod_{\ell=1}^{L_x}\left(\frac1N\sum_{n=1}^Nc_\theta(x_\ell\mid Z_n)\right).
 \label{eq:token-aggregation-mismatch}
\end{equation}
The density ratio also belongs to the complete latent sequence. Independent tokenwise combinations therefore do not estimate the same sequence marginal.

There is a separate algebraic reason an empirical curve can decrease. For any positive weights partitioned into $B$ groups of size $n$, let $a_b$ be the mean weight in group $b$. Then
\begin{equation}
 -\log\left(\frac1B\sum_{b=1}^B a_b\right)
 \leq\frac1B\sum_{b=1}^B[-\log a_b].
 \label{eq:pooled-empirical-monotonicity}
\end{equation}
Thus merging equal-size groups cannot increase their averaged negative-log estimate, even if the weights are numerically inaccurate. This deterministic grouping inequality is distinct from the expected-objective theorem. A sequence formed by adding individual draws need not decrease at every step. Neither observed grouping monotonicity nor a high empirical ESS certifies that the unobserved tail of the importance weights has been explored.

\subsection{Reference adaptations and the sampled Gaussian distance}\label{app:reference-density}
A normalized affine Gaussian reference can be written
\begin{equation}
 Z=\mu_\eta(x)+\sigma_0 B_\eta(x)\epsilon,\qquad
 \epsilon\sim\mathcal N(0,I_D),\qquad \det B_\eta(x)\ne0.
 \label{eq:affine-reference}
\end{equation}
Here $D=L_xd$ is the full latent dimension. Its pointwise log-density is
\begin{align}
 \log r_\eta(z\mid x)
 &=-\frac12\left\|B_\eta^{-1}\frac{z-\mu_\eta(x)}{\sigma_0}\right\|^2
 -\frac D2\log(2\pi)-D\log\sigma_0-\log|\det B_\eta|\notag\\
 &=-\frac12\|\epsilon\|^2-\frac D2\log(2\pi)
 -D\log\sigma_0-\log|\det B_\eta|.
 \label{eq:affine-reference-logdensity}
\end{align}
For covariance $C_\eta=\sigma_0^2B_\eta B_\eta^\top$, the determinant contribution is equivalently $-\tfrac12\log\det C_\eta$. This distinguishes a covariance determinant from that of its square root. The mean family $\mu_\lambda(x)=\alpha_0e(x)+\lambda\sigma_0v(x)$ with $\|v(x)\|^2=D$ and $B_\eta=I_D$ changes only the reference mean; diagonal, grouped, or Cholesky choices of $B_\eta$ change covariance. A zero direction can be assigned no shift. These formulas define valid references without assuming that the model posterior is Gaussian.

For a Gaussian reference conditional on $x$, its mean and covariance must be fixed before drawing the scored $\epsilon$, or fixed conditional on independent auxiliary randomness. Making either parameter a function of that same $\epsilon$ generally creates a different pushforward distribution requiring its own density calculation. Reference adaptations must evaluate the density that actually generated the sample. Freezing $p_\theta$, $c_\theta$, and their boundary conditions keeps the target $P_\theta$ unchanged.

The Gaussian distance in \eqref{eq:affine-reference-logdensity} is a random part of each weight. With exact model density, define $u_\eta(z,x)=\log r_\eta(z\mid x)-\log p_\theta(z)-\log c_\theta(x\mid z)$ and $Q=\|\epsilon\|^2/2$. Replacing $Q$ by $D/2$ produces
\begin{equation}
 u_\eta^{\mathrm{ent}}=u_\eta+Q-\frac D2,\qquad
 \E u_\eta^{\mathrm{ent}}=\E u_\eta=U_{r_\eta}(x).
 \label{eq:entropy-replacement}
\end{equation}
This is a zero-mean control-variate change to a linear expectation, with
\begin{equation}
 \operatorname{Var}(u_\eta^{\mathrm{ent}})-\operatorname{Var}(u_\eta)
 =\operatorname{Var}(Q)+2\operatorname{Cov}(u_\eta,Q).
 \label{eq:entropy-replacement-variance}
\end{equation}
The change need not reduce variance. It cancels from paired reference differences using the same $\epsilon$ and dimension. However, exponentiating the modified expression gives
\begin{equation}
 e^{-u_\eta^{\mathrm{ent}}}
 =w_{r_\eta}(z,x)\exp\left(\frac D2-Q\right),
 \label{eq:entropy-replacement-weight}
\end{equation}
whose expectation is not generally $P_\theta(x)$. Even its additional factor is not mean one: $\E_\epsilon e^{D/2-Q}=(e/2)^{D/2}$, and it is generally correlated with $w_{r_\eta}$. Therefore the exact sampled distance must be retained when forming importance weights. Integrating the entire weight analytically would be a different, potentially valid operation; replacing one term in its logarithm by its expectation is not that operation.

\subsection{Conditional density noise and apparent gap recovery}\label{app:noisy-weights}
For exact weights, the sample-average likelihood is unbiased and the negative logarithm introduces the nonnegative gap. Noise in the logarithm changes the likelihood estimator before this last operation. To isolate that effect, let $\widehat\ell_n=\ell_n+\xi_n$ with $\E[\xi_n\mid Z_{1:N},x]=0$. The function
\[
 F(\ell_{1:N})=-\log\left(N^{-1}\sum_ne^{\ell_n}\right)
\]
is concave. Conditional Jensen therefore implies
\begin{equation}
 \E[F(\ell_{1:N}+\xi_{1:N})\mid Z_{1:N},x]
 \leq F(\ell_{1:N}).
 \label{eq:conditional-lognoise-shift}
\end{equation}
For $N=1$, $F$ is linear and the shift is zero. For $N>1$, centered log-weight noise can lower the expected objective. This statement allows correlated errors under the stated conditional mean assumption; shared errors need not yield a strict decrease. Biased log-density errors have no such general sign guarantee.

\begin{proof}[Proof of Proposition~\ref{prop:noisy-tightening}]
At the exact posterior, $w_r(Z_n,x)=P_\theta(x)$. Hence the noisy estimate is $-\log P_\theta(x)-\log(N^{-1}\sum_ne^{\xi_n})$. Independent Gaussian noise gives $\E e^{\xi_n}=e^{v/2}$, and the strong law proves \eqref{eq:noisy-zero-gap}. The leave-one-out argument in \eqref{eq:iwae-monotonic-proof} applies to the positive independent variables $e^{\xi_n}$, proving monotonicity of the expectation. At $N=1$, the error has zero mean. The variance $v$ is in squared nats per sequence, yielding the stated offset after dividing by $L_x$.
\end{proof}

More generally, suppose the pairs $(Z_n,\xi_n)$ are independent and identically distributed and have finite positive mean weight. Write $m_r(z,x)=\E[e^\xi\mid z,x]$. Then
\begin{align}
 \E[e^{\widehat\ell_r}]&=P_\theta(x)
 \E_{Z\sim p_\theta(\cdot\mid x)}m_r(Z,x),\notag\\
 -\log\left(\frac1N\sum_ne^{\widehat\ell_{r,n}}\right)
 &\xrightarrow[N\to\infty]{\mathrm{a.s.}}-\log P_\theta(x)
 -\log\E_{p_\theta(\cdot\mid x)}m_r(Z,x).
 \label{eq:noisy-weight-limit}
\end{align}
The limiting offset depends on a posterior-weighted exponential moment, not merely an average log-variance under the reference. Correlation across draws, state-dependent variance, and solver bias require their actual error model. In particular, a Gaussian $v/2$ correction is not justified by a measured variance alone.

Holding the numerical state trajectory and grid fixed, a Hutchinson estimator uses independent zero-mean probes $V_m$ with $\E V_mV_m^\top=I_D$:
\begin{equation}
 \widehat{\operatorname{tr}}_M(J)
 =\frac1M\sum_{m=1}^M V_m^\top J V_m,
 \qquad \E\widehat{\operatorname{tr}}_M(J)=\operatorname{tr}J.
 \label{eq:hutchinson-trace}
\end{equation}
For a deterministic quadrature rule independent of these probes, the resulting integrated log-density estimate is unbiased for that quadrature target \citep{grathwohl2019ffjord}. If independent probe trajectories are averaged, its conditional trace variance is $v(z)/M$. Reusing one probe across time changes temporal covariances; resampling at each node changes them again. Neither removes state-integration bias. Probe-dependent adaptive grids require separate analysis because the fixed-grid conditioning argument then need not apply.

A fixed-latent diagnostic repeats density evaluation for one $z$. Any decrease obtained by exponentiating and pooling these repeats occurs without increasing latent coverage and isolates an effect of density noise under those numerical settings. Paired comparisons of references can nevertheless have much smaller variance than either absolute estimate, since
\begin{equation}
 \operatorname{Var}(\widehat u_a-\widehat u_b)
 =\operatorname{Var}(\widehat u_a)+\operatorname{Var}(\widehat u_b)
 -2\operatorname{Cov}(\widehat u_a,\widehat u_b).
 \label{eq:paired-estimator-variance}
\end{equation}
Absolute trace variance alone therefore does not determine the uncertainty of a paired reference improvement.

For bookkeeping, the token-normalized multi-sample report can be decomposed as
\begin{equation}
 \widehat{\mathcal U}_{r,N}
 =\mathcal L_\theta+\mathcal G_{r,N}
 +b_{\mathrm{impl}}+b_{\mathrm{num}}+b_{\mathrm{lognoise}}+\eta_{\mathrm{MC}}.
 \label{eq:multisample-error-decomposition}
\end{equation}
To avoid double counting, define these terms sequentially: implementation changes the specified ideal objective; numerical error changes it to the expectation obtained with deterministic numerical log-weights; $b_{\mathrm{lognoise}}$ is the additional expected change from random log-weights; and $\eta_{\mathrm{MC}}$ is centered fluctuation around that final procedure. At $N=1$, centered additive log-weight noise contributes no $b_{\mathrm{lognoise}}$. At larger $N$, \eqref{eq:conditional-lognoise-shift} explains why this term can be negative. Confidence intervals for $\eta_{\mathrm{MC}}$ do not bound the other terms.

\subsection{Finite solver maps and continuous-flow density}\label{app:solver-density}
Unbiased divergence estimation does not remove the discrepancy between a numerical map and the continuous flow it approximates. For the one-dimensional ODE $\dot y=y$, one Heun step of length $h$ gives
\begin{equation}
 y_{n+1}=(1+h+h^2/2)y_n,\qquad
 \log\left|\frac{\partial y_{n+1}}{\partial y_n}\right|
 =\log|1+h+h^2/2|.
 \label{eq:heun-map-jacobian}
\end{equation}
The integral of the divergence $1$ is $h$, which is generally different. Thus Heun integration of states and divergence approximates the continuous-flow density; it does not automatically compute the exact density of the finite-step map. Increasing $M$ can reduce trace variation while leaving this discrepancy unchanged. Increasing $S$ and checking finite precision address other errors, and agreement at two settings is not a rigorous bias bound. At finite terminal noise, replacing a non-Gaussian terminal marginal by a Gaussian prior can also change the model itself; a density check must use a reference likelihood with the same prior and decoder.

\subsection{A sandwich identity on model-generated texts}\label{app:model-sandwich}
For a fixed text let $a_x(z)=\log r(z\mid x)-\log p_\theta(x,z)=-\log w_r(z,x)$. Assuming both directional KL divergences are finite,
\begin{align}
 U_r(x)&=\E_{r(z\mid x)}a_x(Z)
 =-\log P_\theta(x)+\KL(r\Vert p_\theta(\cdot\mid x)),\notag\\
 B_r(x)&=\E_{p_\theta(z\mid x)}a_x(Z)
 =-\log P_\theta(x)-\KL(p_\theta(\cdot\mid x)\Vert r),\notag\\
 U_r(x)-B_r(x)&=\KL(r\Vert p_\theta(\cdot\mid x))
 +\KL(p_\theta(\cdot\mid x)\Vert r).
 \label{eq:model-sandwich}
\end{align}
Hence $B_r(x)\leq-\log P_\theta(x)\leq U_r(x)$ in ideal expectation. An exact ancestral sample $(X,Z)\sim p_\theta(z)c_\theta(x\mid z)$ supplies a posterior draw conditional on the sampled $X$. Averaging such pairs can estimate the model-text average of $B_r$; it does not supply arbitrarily many posterior draws for a chosen held-out text. Reference draws conditional on those same generated texts estimate the corresponding average of $U_r$.

This use of simulated data is related to bidirectional Monte Carlo validation \citep{grosse2015bdmc}; \eqref{eq:model-sandwich} is a direct reference-distribution identity, not a claim to implement bidirectional annealed importance sampling. Its validity requires a joint sampler and density for the same model, including sampling from the categorical decoder rather than replacing it with argmax. Finite-step flow sampling must be reconciled with the density issue in Appendix~\ref{app:solver-density}. Finally, these averages concern $X\sim P_\theta$, whereas validation likelihood concerns $X\sim P$. Neither the gap nor its sandwich width transfers between these text distributions without further assumptions. Individual Monte Carlo realizations need not provide an upper and lower bracket.

\subsection{A normalized construction for ranking ambiguity}\label{app:ranking-proof}
\begin{proof}[Proof of Proposition~\ref{prop:ranking}]
Consider two possible observed texts, $x^\star$ and $x^\circ$, and an evaluation set containing $x^\star$. For any $a>0$ and $g\geq0$, define
\[
 P_\theta(x^\star)=e^{-a},\qquad
 P_\theta(x^\circ)=1-e^{-a}.
\]
Let $z$ be one-dimensional and choose conditional latent distributions
\[
 p_\theta(z\mid x^\star)=\mathcal N(\sqrt{2g},1),
 \qquad p_\theta(z\mid x^\circ)=\mathcal N(0,1).
\]
These specify a normalized joint distribution. Its latent marginal and decoder can be obtained by marginalization and Bayes' rule, giving a valid model of the form \eqref{eq:discrete-marginal}. Use the same reference $r(z\mid x^\star)=\mathcal N(0,1)$ for every construction. At $x^\star$, exact NLL is $a$, posterior KL is $g$, and the bound is $a+g$.

For the first pair of models, set $(a_A,g_A)=(1,1)$ and $(a_B,g_B)=(2.5,0.5)$. For the second pair, set $(a_A,g_A)=(1.9,0.1)$ and $(a_B,g_B)=(0.5,2.5)$. Both pairs report $U_A=2$ and $U_B=3$. In the first pair $a_A<a_B$, while in the second $a_A>a_B$. All distributions are normalized and all gaps are nonnegative. A common token normalization preserves the rankings.
\end{proof}

The construction proves that bound values alone are insufficient; it makes no assumption that a particular trained CDLM realizes these exact posteriors. Additional structural knowledge or a separate estimate of the gaps can supply information that is absent from the reported pair of numbers.

\subsection{What a common scale reveals about covariance}\label{app:scale-proof}
For posterior $\mathcal N(\mu+\delta,\Sigma)$ and reference $\mathcal N(\mu,\rho^2\sigma^2I_D)$, Gaussian KL gives
\begin{align}
 G_\rho=\frac12\big[
 &\rho^2\sigma^2\operatorname{tr}(\Sigma^{-1})
 +\delta^\top\Sigma^{-1}\delta-D \notag\\
 &+\log\det\Sigma-D\log\sigma^2-2D\log\rho\big].
 \label{eq:general-gaussian}
\end{align}
The optimal scale satisfies
\begin{equation}
 \rho_*^2=\frac{D}{\sigma^2\operatorname{tr}(\Sigma^{-1})}.
\end{equation}
A relative scale curve can identify this precision-trace combination within the assumed Gaussian family. It cannot identify the mean mismatch, determinant, eigendirections, or full covariance spectrum. In particular, $\sigma^2\operatorname{tr}(\Sigma^{-1})=D$ is compatible with $\rho_*=1$ and substantial remaining gap. Observing a similar curve also does not establish that the true posterior belongs to a Gaussian family.

For a normalized discrete model, $-\log P_\theta(x)\geq0$ gives the trivial inequality $G_r(x)\leq U_r(x)$. The scale-search argument does not remove this inequality; it shows that a relative scale curve or failure to tighten does not generally supply a stronger certificate that the gap is small.

Fitting a quadratic can introduce a further artifact. Even the exact function $384(\rho^2-1-2\log\rho)$ fitted at $\rho\in\{0.98,0.99,1,1.01,1.02\}$ gives an apparent optimum near $1.00005667$ and a fitted improvement of about $2.47\times10^{-6}$ nats/token. This is polynomial approximation error, not a true gain. The archived data's local fitted improvement, $6.31\times10^{-7}$, is smaller still. A positive fitted minimum improvement is therefore not evidence that a measurable nondefault optimum exists.

More generally, a fitted curve $at^2+bt+c$ with $a>0$ has a fitted improvement $b^2/(4a)\geq0$ over $t=0$. A positive lower percentile of this nonnegative fitted statistic is not automatically a valid test of improvement over the null. Finite-grid searches and intervals for individual candidates do not give a uniform bound over all scales without additional assumptions; independent confirmation must evaluate the selected candidate directly.

\section{Archived Likelihood Experiment}\label{app:likelihood-protocol}
\subsection{Estimator and fixed model}
The archived LangFlow calculation uses $\gamma=\log(\sigma^2/\alpha^2)$, $\alpha^2=(1+e^\gamma)^{-1}$, $\sigma^2=(1+e^{-\gamma})^{-1}$, $s=\alpha/\sigma=e^{-\gamma/2}$, and $y=z/\sigma$. The evaluated dynamics and their divergence take the form
\begin{equation}
 \frac{\mathrm dy}{\mathrm ds}=f_\theta(y,s)
 =\widehat e_\theta(\sigma(s)y,\gamma(s)),\qquad
 \nabla_y\!\cdot f_\theta
 =\sigma(s)\operatorname{tr}\frac{\partial\widehat e_\theta}{\partial z}.
 \label{eq:langflow-rescaled-dynamics}
\end{equation}
Integrating from the low-noise endpoint $s_0$ to the high-noise endpoint $s_1<s_0$, define
\[
 A(z_0)=\int_{s_0}^{s_1}\nabla_y\cdot f_\theta(y_s,s)\,\mathrm ds,
 \qquad y_1=\Phi_\theta(z_0/\sigma_0).
\]
With standard Gaussian terminal density on $y_1$, the endpoint log-density is
\begin{equation}
 \log p_\theta(z_0)
 =-\tfrac12\|y_1\|^2-\tfrac D2\log(2\pi)+A(z_0)-D\log\sigma_0.
 \label{eq:langflow-endpoint-density}
\end{equation}
Sampling $z_0=\alpha_0e(x)+\rho\sigma_0\epsilon$ with $\epsilon\sim\mathcal N(0,I_D)$ gives
\begin{align}
 \log r_\rho(z_0\mid x)
 &=-\tfrac12\|\epsilon\|^2-D\log(\rho\sigma_0)-\tfrac D2\log(2\pi),\notag\\
 \Lambda_{r_\rho}(z_0,x)
 &=A(z_0)-\tfrac12\|y_1\|^2+\tfrac12\|\epsilon\|^2+D\log\rho.
 \label{eq:langflow-pointwise-ratio}
\end{align}
The Gaussian normalization constants cancel, but the sampled quadratic terms remain. Substituting this ratio into $u_\rho=-\log c_\theta(x\mid z_0)-\Lambda_{r_\rho}(z_0,x)$ yields
\begin{equation}
 u_\rho=-\log c_\theta(x\mid z_0)-A(z_0)
       +\tfrac12\|y_1\|^2-\tfrac12\|\epsilon\|^2-D\log\rho.
 \label{eq:archived-estimator}
\end{equation}
Numerical integration and stochastic divergence estimation produce the recorded approximation to this expression. Omitting $-D\log\rho$ would not compare the valid reference family in \eqref{eq:scale-family}.

The checkpoint SHA-256 prefix is \texttt{9dcc2573297d9f0a}. The endpoint values are $\gamma_0=2.60142684$ and $\gamma_1=16.05187035$, with $\alpha_0=0.26276729$ and $\sigma_0=0.96485919$. The main run uses an A100-SXM4-80GB GPU and batch size 16. State, grid, and reconstruction cross-entropy use float32; selected accumulation and density combinations use float64. The split seed is 20261102. Every reference comparison freezes the weights, embedding, decoder, endpoints, velocity field, and terminal prior.

\subsection{Matched numerical controls}
\begin{table}[htbp]
 \centering\small
 \caption{\textbf{Matched numerical controls.} Eight rows and two shared draws per row. The change is the new setting's $\Delta$ minus the 128-step/one-probe $\Delta$ on those same draws. Errors are standard errors across eight row means, not 95\% intervals. Units are nats/token.}
 \begin{tabularx}{\linewidth}{@{}llrrY@{}}
\toprule
Steps / probes & $\rho$ & Baseline $\Delta$ & New $\Delta$ & Change $\pm$ SE \\
\midrule
128 / 4 & 0.99 & $-0.077967$ & $-0.077911$ & $+0.000056\pm0.001056$ \\
128 / 4 & 0.9998 & $-0.000246$ & $-0.000208$ & $+0.000038\pm0.000651$ \\
256 / 1 & 0.99 & $-0.077967$ & $-0.079265$ & $-0.001298\pm0.001770$ \\
256 / 1 & 0.9998 & $-0.000246$ & $-0.000942$ & $-0.000696\pm0.000525$ \\
\bottomrule
\end{tabularx}

 \label{tab:numerics}
\end{table}
Table~\ref{tab:numerics} avoids comparing an expanded baseline sample with a smaller numerical-control sample. It supports the direction of the larger $\rho=0.99$ deterioration, while leaving micro-scale improvements unresolved. The checks do not separately identify stochastic trace variance and deterministic solver bias.

Replacing $\|\epsilon\|^2/2$ in \eqref{eq:archived-estimator} by its expectation $D/2$ preserves the ideal expected bound but changes variance. In the archived data, retaining the random quadratic yields within-row residual standard deviations $0.03618$ and $0.03642$ nats/token on calibration and evaluation, respectively; replacing it gives $0.50298$ and $0.53743$. These correspond to variance ratios of about $193$ and $218$, not standard-deviation ratios of that size. Correlation with the terminal quadratic explains why analytically averaging one term need not reduce total variance. When the same $\epsilon$ is shared between scales, the two forms give identical paired scale differences.

This replacement is restricted to estimating the single-sample expectation. In the multi-sample expression \eqref{eq:multisample-bound}, each $\Lambda_{r_\rho}$ must instead use the sampled $\|\epsilon\|^2/2$ in \eqref{eq:langflow-pointwise-ratio}, as shown by \eqref{eq:entropy-replacement-weight}.

\subsection{Scope of verification}
The original audit checked 1,792 calibration records and 4,608 held-out records, their split identities, and the reconstruction of \eqref{eq:archived-estimator}. The maximum reported arithmetic reconstruction error was below $3\times10^{-11}$ nats/sequence. This verifies consistency of stored components, not the accuracy of the ODE solution.

These results are based on archived measurements rather than a new checkpoint evaluation. Incomplete experiment drivers and a discrepancy between the archived test source and its recorded version limit exact reproduction of the original run.

\section{Generation-Metric Proofs and Estimator Behavior}
\subsection{Matching two moments with three texts}\label{app:moment-proof}
\begin{proof}[Proof of Proposition~\ref{prop:small-support}]
For texts $x_1,\ldots,x_N$ with common scored length $T>0$, write
\[
 f_i=(\ell_J(x_i)/T,\ H(\widehat p_{x_i}))\in\mathbb R^2.
\]
For any probabilities $w_i$, the score pair before exponentiating its first coordinate is $\sum_iw_if_i$. Suppose more than three weights are positive. The vectors $(1,f_i)\in\mathbb R^3$ on that support are linearly dependent, so there are coefficients $c_i$, not all zero, with $\sum_i c_i=0$ and $\sum_i c_if_i=0$. There must be both positive and negative coefficients. Let
\[
 \tau=\min_{i:c_i>0}\frac{w_i}{c_i},
 \qquad w'_i=w_i-\tau c_i.
\]
All new weights are nonnegative, sum to one, and preserve $\sum_iw_if_i$; at least one positive weight becomes zero. Repeating removes support points until at most three remain. The mean judge NLL and mean unigram entropy are unchanged, hence so is population Gen PPL after exponentiation.
\end{proof}
Arbitrary weights need not be representable exactly by a finite equally weighted sample file, but rational approximations produce arbitrarily close score pairs. This distinction does not affect the population non-identifiability result. For variable scored lengths, matching the relevant numerator and denominator requires additional moments; the stated three-point result is restricted to common length.

\subsection{Sparse contamination and missing modes}
\phantomsection\label{app:mmd-limits}
Let $\mu_P=\E_{X\sim P}k(X,\cdot)$ be a kernel mean. For $Q_\epsilon=(1-\epsilon)P+\epsilon R$,
\[
 \mu_{Q_\epsilon}-\mu_P=\epsilon(\mu_R-\mu_P),
\]
which proves \eqref{eq:contamination} by taking the squared RKHS norm. With an RBF kernel, each within-distribution expectation is at most one and the cross expectation is nonnegative, giving $D_k^2(P,R)\leq2$.

For a missing-mode example, let $A$ and $B$ have disjoint supports and let $P=(1-\epsilon)A+\epsilon B$ while $Q=A$. Then
\[
 D_k^2(P,Q)=\epsilon^2D_k^2(B,A)\leq2\epsilon^2,
 \qquad \KL(P\Vert Q)=+\infty
\]
for every $\epsilon>0$. Small MMD thus provides no general upper bound on forward KL or exact PPL. Sampling itself also limits rare-error detection: if an event has probability $0.001$, its probability of being absent from 512 independent outputs is $(1-0.001)^{512}\simeq0.599$. No event detector can count an error that did not occur in the evaluated sample.

\subsection{Copying a sample file and replaying the reference}\label{app:replication}
Let $S$ contain $r\geq2$ rows, assume $k(x,x)=1$, and let $\bar k_S$ be its mean off-diagonal similarity. Form $S^{\times d}$ by repeating each row $d$ times. The reference term and reference--sample cross term in \eqref{eq:mmd-u} do not change. Counting repeated pairs in the sample term yields
\begin{equation}
 \widehat D_u^2(X,S^{\times d})-\widehat D_u^2(X,S)
 =\frac{(d-1)(1-\bar k_S)}{dr-1}.
 \label{eq:duplication}
\end{equation}
The empirical probability measure of $S$ is unchanged by this copying. Its V-statistic discrepancy, which retains diagonal terms and uses squared sample sizes in the within-set denominators, is therefore invariant. The change in \eqref{eq:duplication} is a finite-sample correction, not proof that information about the underlying generator increased.

In Figure~\ref{fig:support}, the output count is fixed while its support shrinks, which changes the empirical distribution.

If the same $n$-row reference is supplied as both arguments, then
\begin{equation}
 \widehat D_u^2(X,X)=-\frac{2(1-\bar k_X)}{n}.
 \label{eq:reference-replay}
\end{equation}
The two inputs are no longer independent. A negative value in this construction does not contradict the fixed-kernel unbiasedness result under independent sampling, and it is not evidence of higher-than-real quality.

\section{Generation Protocol and Supplementary Results}\label{app:generation-protocol}
\subsection{Data construction and preprocessing}
The OWT pilot's source-row seed is 20261105. It selects 8,192 distinct source-row indices: 512 in a calibration set, and three groups of 512 reference plus 2,048 candidate rows. Each natural control is the first 512 rows in its candidate pool. The donor for splicing is the next 512 rows. Thirteen tokens are replaced beginning at source-token offsets 1, 58, or 114 (zero-based), leaving the first and last tokens untouched. Block reordering maps four consecutive 32-token blocks to the order $(3,1,4,2)$. Sparse mixtures replace the first $\operatorname{round}(512p)$ control rows at fraction $p$.

The selected low-NLL text is drawn from each 2,048-row pool after requiring entropy at least the mean of its control. The easiest quartile uses judge NLL per scored token. The feature-neighbor quartile uses cosine similarity to the first independently selected calibration feature; it is a selection along a feature direction, not a set defined by annotated topic labels. Near-duplicates repeat 64 control texts with distinct numeric suffixes.

For judge likelihood, texts are tokenized by GPT-2, padded with a mask, and scored from the second token through the last valid token. NLL is accumulated in float32 in the audit path; token counts exclude padding and the unscored first token. The 1024-token maximum is explicit. The FP32 feature-swap conditions use matched special-token handling. The MAUVE comparison uses \texttt{mauve-text 0.4.0}, clustering seeds 25, 26, and 27, and the package's standard PCA and clustering settings.

The feature-extraction runs used two A100-SXM4-80GB GPUs, PyTorch 2.3.1, and CUDA 12.1, with TF32 disabled. Weight hashes were checked and recorded. The audit's saved verification reports equality of last-token features and per-row NLL between its optimized extraction path and the repository interface on 16 texts. It also records an exact truncation collision for inputs with identical retained prefixes. These checks validate specified computation paths, not human validity of every constructed text.

\subsection{Representation and clustering ablation}\label{app:extra-results}
\begin{table}[htbp]
 \centering\small
 \caption{\textbf{MAUVE under matched FP32 preprocessing, fold 0.} The same text sets are compared at 32 clusters; entries average three clustering seeds. Feature-neighbor selection is defined using mean features and is therefore a directed diagnostic.}
 \begin{tabularx}{\linewidth}{@{}Yrr@{}}
\toprule
Construction & Mean pooling & Last-token pooling \\
\midrule
Real control & 0.98740 & 0.98441 \\
Judge-easiest quartile & 0.88069 & 0.95798 \\
Feature-neighbor quartile & 0.42981 & 0.95648 \\
Middle splice & 0.97886 & 0.98480 \\
\bottomrule
\end{tabularx}

 \label{tab:feature-ablation}
\end{table}
Table~\ref{tab:feature-ablation} controls precision and preprocessing while varying pooling. Cluster count is also influential: on the last-feature control, 16, 32, and 64 clusters give MAUVE $0.99531$, $0.98441$, and $0.93096$, respectively. For 64 texts repeated to 512 rows, the corresponding values are $0.68034$, $0.20435$, and $0.01815$. These are small-sample diagnostics, not a basis for choosing the most favorable cluster count after observing a result.

\paragraph{Numerical consistency.}
The separate template sets each contain 512 texts. A reference and a control use equal counts in both mentions; a contradictory edit increments only the second count, while a consistent edit increments both. The latter preserves consistency but changes the numerical facts, so it is a consistency-preserving control rather than a claim of complete semantic invariance. Gen PPL increases from $24.560$ for the control to $27.117$.

\paragraph{Cost.}
With cached features and $n=512$, the median recorded MAUVE time for three clustering seeds is about $7.18$ seconds. The audit's multi-view feature and likelihood extraction for 512 ordinary texts takes roughly 15--17 seconds; doubled spaces increase tokenization length and extraction time. The benchmark must report these costs separately and match representation and repetition settings when comparing metric efficiency.

\end{document}